\documentclass[letterpaper]{article} % DO NOT CHANGE THIS
\usepackage{aaai23}  % DO NOT CHANGE THIS
\usepackage{times}  % DO NOT CHANGE THIS
\usepackage{helvet}  % DO NOT CHANGE THIS
\usepackage{courier}  % DO NOT CHANGE THIS
\usepackage[hyphens]{url}  % DO NOT CHANGE THIS
\usepackage{graphicx} % DO NOT CHANGE THIS
\usepackage{natbib}  % DO NOT CHANGE THIS AND DO NOT ADD ANY OPTIONS TO IT
\usepackage{caption} % DO NOT CHANGE THIS AND DO NOT ADD ANY OPTIONS TO IT
\usepackage{amsfonts}       % blackboard math symbols
\usepackage{nicefrac}       % compact symbols for 1/2, etc.
\usepackage{microtype}      % microtypography
\usepackage{xcolor}         % colors
\usepackage{colortbl}
\usepackage{graphicx,setspace,latexsym,amsmath,amssymb,amsthm,bm}
\usepackage[font=small,labelfont=bf]{caption}
\usepackage{subcaption}
\usepackage{import}
\usepackage{dirtytalk}
\usepackage{algorithm,algorithmic}
\usepackage{subcaption}
\usepackage{changepage}
\usepackage{enumitem}
\usepackage[affil-it]{authblk} 
\usepackage{etoolbox}
\usepackage{lmodern}
\usepackage{tikz}
\usepackage[draft]{commands}
\usepackage{minitoc}

\renewcommand{\ind}{\mathbf{1}}
\usepackage{newfloat}
\usepackage{listings}
\floatstyle{ruled}
\newfloat{listing}{tb}{lst}{}
\floatname{listing}{Listing}
\title{Efficient and Accurate Learning of Mixtures of Plackett-Luce Models}
\author {
    Duc Nguyen\textsuperscript{\rm 1},
    Anderson Y. Zhang\textsuperscript{\rm 2}
}

\affiliations {
    \textsuperscript{\rm 1} Depart of Computer and Information Science, University of Pennsylvania \\
    \textsuperscript{\rm 2} Department of Statistics and Data Science, University of Pennsylvania \\
    mdnguyen@seas.upenn.edu, ayz@wharton.upenn.edu
}

\newcommand{\btheta}{\bm{\theta}}

\newcommand{\cml}{\text{CML}}

\newcommand{\lsr}{\text{LSR}}
\newcommand{\gmm}{\text{GMM}}

\usepackage{titletoc}
\usepackage[page,header]{appendix}

\colorlet{bestcolor}{gray!30}
\newcommand\shade{\cellcolor{bestcolor}}

\begin{document}

\maketitle
\doparttoc % Tell to minitoc to generate a toc for the parts
\faketableofcontents % Run a fake tableofcontents command for the partocs

\part{} 
\parttoc % Insert the document TOC

\begin{abstract}
Mixture models of Plackett-Luce (PL) -- one of the most fundamental ranking models -- are an active research area of both theoretical and practical significance. Most previously proposed parameter estimation algorithms instantiate the EM algorithm, often with random initialization. However, such an initialization scheme may not yield a good initial estimate and the algorithms require multiple restarts, incurring a large time complexity. As for the EM procedure, while the E-step can be performed efficiently, maximizing the log-likelihood in the M-step is difficult due to the combinatorial nature of the PL likelihood function (Gormley and Murphy 2008). Therefore, previous authors favor algorithms that maximize surrogate likelihood functions (Zhao et al. 2018, 2020). However, the final estimate may deviate from the true maximum likelihood estimate as a consequence. In this paper, we address these known limitations. We propose an initialization algorithm that can provide a provably accurate initial estimate and an EM algorithm that maximizes the true log-likelihood function efficiently. Experiments on both synthetic and real datasets show that our algorithm is competitive in terms of accuracy and speed to baseline algorithms, especially on datasets with a large number of items.
\end{abstract}

\section{Introduction}

Learning to rank is an active area of research with wide-ranging applications in recommendation systems, information retrieval, crowdsourcing and the social sciences.
The Plackett-Luce (PL) model \cite{plackett1975analysis,luce2012individual} is one of the most fundamental ranking models.
In a universe of $n$ items, the PL model posits that item $i$ has a latent \emph{utility} $\theta^*_i\in \R$. The probability of observing a full ranking $\pi$ given by the user (most preferred item first) is given as
\begin{multline}\label{eqn:plackett-luce-likelihood}
\Pr^{PL}(\pi= [\pi_1,.., \pi_n]\,\lvert \theta^*) = \prod_{i=1}^{n-1} \frac{\exp{\theta^*_{\pi_i}}}{\sum_{j=i}^n \exp{\theta^*_{\pi_j}} }\,.
%  \frac{e^{\theta^*_{\pi_{1}}}}{\sum_{j=1}^n e^{\theta^*_{\pi_{j}}}} \ldots \frac{e^{\theta^*_{\pi_{n-1}}}}{ e^{\theta^*_{\pi_{n-1}}} + e^{\theta^*_{\pi_{n}}} } \,.
\end{multline}
The maximum likelihood estimate (MLE) can be obtained using iterative algorithms such as the Minorize-Maximize (MM) algorithm of \citet{hunter2004mm} and enjoys favorable theoretical properties \cite{hajek2014minimax}. In recent years, an algorithm known as Luce spectral ranking (LSR) \cite{maystre2015fast} has become the method of choice for maximum likelihood inference for PL models. LSR outputs the MLE just like MM but is often much faster. 

The PL model is closely connected to the Bradley-Terry-Luce (BTL) model \cite{luce2012individual} for \emph{pairwise comparisons}. For two items $i\neq j$, the probability that $i$ is ranked ahead of $j$ \emph{in a ranking} is equal to the probability that $i$ beats $j$ in a \emph{pairwise comparison} under the BTL model. That is,
\begin{equation}\label{eqn:btl}
 \Pr^{PL}(\pi(i) < \pi(j)) = \Pr^{BTL}_{ij} = \frac{1}{1 + \exp{-(\theta^*_i - \theta^*_j)}}\,, 
\end{equation}
where $\pi(i)$ is the position of item $i$ in ranking $\pi$.

The classical PL model assumes that there is a universal preference ordering of the items according to their utilities.
However, in practice, there might be multiple subpopulations of users with different preference profiles which cannot be fully captured by a single PL model. In such settings, a mixture of PL models is a more appropriate modeling assumption.

\textbf{Problem Descriptions.} Consider a mixture model with $K$ components and $n$ items for some constant $K$. Let $\beta^* = [\beta^*_1,\ldots, \beta^*_K]^\top $, ${\beta^*}^\top \mb 1 = 1$ denote the mixing distribution. For component $k \in [K]$ (where $[a]$ denotes $[1,\ldots, a]$), the utility parameters for the items are
$$ \theta^{*k} = [\theta^{*k}_1,\ldots, \theta^{*k}_n]^\top \in \R^n\,. $$
Let $\btheta^* = [\theta^{*1}, \ldots, \theta^{*K}] \in \R^{n \times K}$ denote the concatenation of the $K$ sets of parameters. A \emph{ranking dataset} $\Pi$ is a collection of full rankings. 

Consider the following generative model for a ranking dataset of size $m$.
For $l \in [m]$, let ${z^*_l} \in [K]$ denote the mixture component membership where $\Pr({z^*_l} = k) = \beta^*_k$. Then a permutation $\pi_l$ is drawn from the PL distribution parametrized by $\theta^{{*z_l}}$. That is,
\begin{equation}\label{eqn:plackett-luce-likelihood-mixtures}
\Pr^{PL}(\pi_l = [\pi_{l,1},\ldots,\pi_{l,n}] \,\lvert\, {z^*_l}, \btheta^*) = \prod_{i=1}^{n-1} \frac{\exp{\theta^{*z^*_l}_{\pi_{l,i}}}}{\sum_{j=i}^n \exp{\theta^{*z_l^*}_{\pi_{l,j}}} } \,,
% \frac{e^{\theta^{*{z^*_l}}_{\pi_{l,1}}}}{\sum_{j=1}^n e^{\theta^{*{z^*_l}}_{\pi_{l,j}}}} \cdot \ldots \cdot \frac{e^{\theta^{*{z^*_l}}_{\pi_{l,n-1}}}}{ e^{\theta^{*{z^*_l}}_{\pi_{l,n-1}}} + e^{\theta^{*{z^*_l}}_{\pi_{l,n}}} } \,,
\end{equation}
where $\pi_{l,i}$ denote the $i$-th item in permutation $\pi_l$. The reader may recognize two identifiability issues here. The first is parameter translation. For each component, the distributions parametrized by $\theta^{*k}$ and $\theta^{*k} + c\cdot \mb 1_n$ are the same for any $c\in \R$. The second is mixture components (columns of $\btheta^*$) relabeling. To account for these issues, we consider the following error metric.
\begin{equation}\label{eqn:metric} \text{dist}(\btheta, \btheta^*) := \min_{R \in \mathcal{O}^{K\times K}} \lVert N(\btheta) R - N(\btheta^*) \rVert_F \,,
\end{equation}
where $\mathcal{O}^{K\times K}$ is the set of all permutation matrices \cite[Chapter 2]{strang1993introduction} of size $K\times K$ and $N$ is the normalization operator (i.e., $N(\btheta)_{\cdot\, k} = \theta^k - \frac{1}{n} (\theta^k)^\top \mb 1_n $). 

\textbf{Prior Works.} Generalizing the PL model to mixtures adds a layer of complexity to the inference problem. In general, the likelihood function is non-convex in the model parameters. Most previously proposed algorithms instantiate the EM algorithm \cite{dempster1977maximum}. As a general recipe, an EM algorithm is initialized with some parameter $\btheta^{(0)}$ (e.g., using random initialization). It then repeats the following two steps for $t=1,2,\ldots$ until convergence.

The E-step computes the posterior class probability conditioned on the current estimate:
\begin{equation}\label{eqn:posterior}
q_l^k =  \Pr({z^*_l} = k \,\lvert \, \pi_l, \btheta^{(t-1)}) \propto \beta_k \cdot \Pr^{PL} (\pi_l \,\lvert {z^*_l} = k, \btheta^{(t-1)})\, 
\end{equation}
for $l\in[m], k\in [K]$ where $\Pr^{PL}$ is given in Equation (\ref{eqn:plackett-luce-likelihood-mixtures}) and $\beta$ the prior class probability. Thanks to the closed form of the PL likelihood function, the E-step can be done efficiently. The M-step obtains the next estimate $\btheta^{(t)}$ by maximizing the joint log-likelihood function which decomposes into $K$ \emph{weighted log-likelihood functions}. Namely,
\begin{equation}\label{eqn:weighted-log-likelihood}
\btheta^{(t)} = \arg\max_{\btheta}\sum_{k=1}^K \bigg(\sum_{l=1}^m q_l^k \log \Pr^{PL}(\pi_l, {z^*_l} = k \,\lvert\, \btheta)\bigg) \,. 
\end{equation}
% The functions $\{\mathcal{L}(\Pi, q^k, \btheta)\}_{k=1}^K$ are \emph{weighted log-likelihood functions}. 
Due to the combinatorial nature of the PL likelihood function, the derivative of the log likelihoood function has a complicated form. As a result, maximizing the (weighted) log-likelihood via gradient-based algorithms quickly becomes inefficient as $n$ grows.

The first practical approach towards solving the M-step uses the Minorize-Maximize algorithm of \citet{hunter2004mm}, yielding the EMM algorithm of \citet{gormley2008exploring}. While guaranteed to solve the M-step, it has been observed that the MM subroutine converges slowly even for datasets with a moderate number of items (e.g., Figure \ref{fig:ml-10M}). Motivated by practical concerns, researchers have developed \emph{pseudo-likelihood} estimators that optimize, instead of the true log-likelihood function, \emph{alternative objective functions}. Two such algorithms are the Generalized Method of Moments (GMM) of \citet{azari2013generalized} and Composite Marginal Likelihood (CML) of \citet{zhao2018composite}. It has been observed experimentally that GMM is considerably faster than MM and CML is even faster than GMM with comparable accuracy. Besides maximum likelihood (ML) inference methods, previous authors have also proposed Bayesian inference algorithms \cite{guiver2009bayesian,mollica2017bayesian}. In this paper, we focus primarily on ML algorithms but include additional experiments with Bayesian methods in the supplementary materials. 

Using GMM and CML to solve the M-step gives us the EM-GMM algorithm \cite{zhao2018learning} and the EM-CML algorithm \cite{zhao2020learning}, respectively. The only non-EM algorithm for learning PL mixtures that we are aware of is a GMM-based algorithm proposed in \citet{zhao2016learning,zhao2019learning}. However, the construction of the algorithm is quite ad-hoc and the authors did not show extension of the algorithm to more than 2 mixture components. In addition, previous authors primarily restrict their experiments to datasets with a small number of items such as the SUSHI datasets \cite{kamishima2003nantonac} with $n=10$. It is unknown how the previous methods perform when $n$ is large. Recent works have also studied PL mixtures learning with features and partial rankings \citep{tkachenko2016plackett,liu2019learning}. While we include possible extensions of our algorithm in the supplementary materials, our main focus in this paper is an \emph{improved algorithm for the classical setting}.

\textbf{Our Contributions.} We propose a new EM algorithm for learning mixtures of PL models that
\begin{itemize}
\item Has a provably accurate initialization procedure with a finite sample error guarantee, the first of its kind in the literature;
\item Efficiently maximizes the weighted log-likelihood function in the M-step without using a surrogate likelihood or objective function, thus returning the true maximum likelihood estimate;
\item Performs competitively with the previously proposed algorithms in terms of accuracy and speed, and is scalable to datasets with $n \geq 100$.
\end{itemize}

\section{The Spectral EM Algorithm}\label{sect:algorithm}

In this section, we present our algorithmic contributions. Section \ref{subsect:spectral-initialization} describes the spectral initialization algorithm and Section \ref{subsect:em-refinement} describes the EM refinement procedure.

\subsection{Spectral Initialization}\label{subsect:spectral-initialization}
The initialization for our algorithm is delegated to spectral clustering (Algorithm \ref{alg:spectral-clustering}) and a least squares minimization algorithm (Algorithm \ref{alg:least-squares-estimate}). To apply spectral clustering, we first embed each ranking $\pi_l$ into a `pairwise vector' -- $X_l \in \{0,1\}^{{n\choose 2}}$ where each entry corresponds to a pair of items. As an overload of notation, we use $d = (d_1, d_2)$ where $d_1 < d_2$ to denote the entry corresponding to the pair $(d_1, d_2)$. Define
\begin{equation}\label{eqn:embedding}
X_{l,d}(\pi) = \begin{cases} 1 &\text{ if } \pi_l(d_1) < \pi_l(d_2) \\ 0 &\text{ otherwise }\end{cases}\,.
\end{equation}
Let $X \in \R^{m \times {n\choose 2}}$ denote the concatenation of the embeddings of $m$ rankings in dataset $\Pi$. Given a target number of components $K$, Algorithm \ref{alg:spectral-clustering} can then be applied to the rows of $X$ to obtain $K$ clusters, $\{\hat C^{k}\}_{k=1}^K \subseteq [m]$.

For each cluster of rankings $\hat C^k$, we estimate the preference probability for a pair $(i, j)$ as 
\begin{equation}\label{eqn:pairwise-preference}
\hat P^k_{ij} =\frac{1}{\lvert \hat C^k\rvert} \sum_{l\in \hat C^k} \ind[\pi_l(i) < \pi_l(j)] \,.
\end{equation}
From the preference probability estimates for all pairs, Algorithm \ref{alg:least-squares-estimate} recovers the utility parameter $\hat\theta^k$. It applies the logit function on the pairwise probabilities and solves a constrained least squares minimization problem, which can be efficiently done using off-the-shelf solvers \cite{scipy}. Algorithm \ref{alg:spectral-init} summarizes the spectral initialization algorithm.
\begin{algorithm}[]
\caption{Spectral Clustering with Adaptive Dimension Reduction}
\hspace*{\algorithmicindent} \textbf{Input: } Dataset $\Pi = \{\pi_1, \ldots, \pi_m\}$, number of mixture components $K$ and threshold $T$. \\
\hspace*{\algorithmicindent} \textbf{Output: } $K$ clusters of rankings. \\
\vspace{-0.75em}
\begin{algorithmic}[1]
    \STATE Embed the rankings as the rows of a matrix $X \in \{0,1\}^{m \times {n\choose 2}}$ according to Equation (\ref{eqn:embedding}).
    \STATE Perform SVD: $ X = U S V^\top $, where the singular values are arranged from largest to smallest.
    \STATE Let $\hat r$ be largest index in $[K]$ such that the difference between the successive singular values is greater than $T$, i.e.,
$ \hat r = \max \{ a \in [K] \,:\, S_{aa} - S_{(a+1) (a+1) } \geq T \}\,.$
    \STATE Run k-means on the rows of $X V_{1:\hat r} $ with $K$ clusters:
    \begin{equation*}\label{eqn:k-means} 
    \big(\hat z, \{\hat c_k\}_{k=1}^K \big) = \underset{\substack{z\in \{1,\ldots,K\}^m \\ \{c_k\} \in \R^{\hat r } }}{\arg\min} \sum_{l=1}^m \lVert V_{1:\hat r}^\top X_l - c_{z_l}\rVert_2^2 \,.
    \end{equation*}
    \STATE Return clusters $\hat C^k = \{l \in [m] \,:\, \hat z_l = k\}$ for $k \in [K]$.
\end{algorithmic}
\label{alg:spectral-clustering}
\end{algorithm}

\begin{algorithm}[]
\caption{Least Squares Parameter Estimation}
\hspace*{\algorithmicindent} \textbf{Input: }Pairwise preference matrix $\hat P \in \R^{n\times n}$. \\
\hspace*{\algorithmicindent} \textbf{Output: }Normalized parameter estimate $\hat\theta$. \\
\vspace{-0.75em}
\begin{algorithmic}[1]
    \STATE Solve the least squares optimization problem\\
    $$ \hat\theta = \arg\max_{\theta: \theta^\top \mb 1 = 0} \sum_{i\neq j} (\hat \phi_{ij} - (\theta_i - \theta_j))^2\,, $$ 
    where $ \hat\phi_{ij} = \ln(\hat P_{ij}/(1-\hat P_{ij}))\,. $
\end{algorithmic}
\label{alg:least-squares-estimate}
\end{algorithm}
% \hspace{-2cm}

\begin{algorithm}[]
\caption{Spectral Initialization}
\hspace*{\algorithmicindent} \textbf{Input: } Dataset $\Pi = \{\pi_1, \ldots, \pi_m\}$, number of mixture components $K$. \\
\hspace*{\algorithmicindent} \textbf{Output: } Parameter estimates for $K$ mixture components $\hat\btheta = [\hat\theta^1,\ldots, \hat\theta^K] \in \R^{n\times K}$. \\
\vspace{-0.75em}
\begin{algorithmic}[1]
    \STATE Run Algorithm \ref{alg:spectral-clustering} on $\Pi$ with $T = \sqrt n\sqrt{m+n}\sqrt{\log n}$ to obtain $K$ clusters $\hat C^1, \ldots,\hat C^K$.
    \STATE Estimate the pairwise preference probabilities $\hat P^k_{ij}$ per Equation (\ref{eqn:pairwise-preference}) for each cluster.
    \STATE Run Algorithm \ref{alg:least-squares-estimate} on $\{\hat P^k\}_{k=1}^K$ and return the parameter estimates for $K$ mixture components.
\end{algorithmic}
\label{alg:spectral-init}
\end{algorithm}
% \footnotetext{The threshold $T$ is selected to satisfy a technical condition in the analysis of spectral clustering. Please refer to our supplementary materials for more details.}

\textbf{Remarks.} The application of spectral clustering to mixtures of PL models has also appeared in a manuscript by \citet{shah2018learning}. There, the authors apply the classical spectral clustering algorithm -- clustering the rows of $XV_{1:K}$ -- and their analysis requires a spectral gap condition which is hard to verify. We use spectral clustering with adaptive dimension reduction and our analysis does not require any spectral gap condition \cite{zhang2022leave}. Furthermore, we focus on \emph{parameter estimation} while \citet{shah2018learning} only focus on clustering, resulting in different theoretical guarantees. The choice of threshold $T$ in Algorithm \ref{alg:spectral-init} is to satisfy a mild technical condition in the analysis of spectral clustering. In our experiments, the performance of the EM algorithm does not seem to critically depend on this threshold.

\textbf{Intuition behind Algorithm \ref{alg:least-squares-estimate}.} Recall the connection between the PL model and the BTL model in Equation (\ref{eqn:btl}). Suppose we observe a large sample drawn from a single PL distribution. Then $\hat P_{ij} \approx P_{ij}^* = 1/\big(\exp{-(\theta_i^*-\theta_j^*)}\big)$ and $ \hat\phi_{ij} = \ln(\hat P_{ij}/1-\hat P_{ij}) \approx \theta^*_i - \theta^*_j $. Solving the least squares optimization problem recovers $\hat\theta \approx \theta^*$. In the mixture setting, if the estimates $\hat P^k$'s are accurate, we obtain good parameter estimates (e.g., Theorem \ref{thm:top-L-mixture-error-init}). \citet{rajkumar2016can} apply a similar idea in their algorithm for \emph{ranking from comparisons of $O(n\log n$) pairs under a single BTL model}. They first apply the logit function on the pairwise preference probabilities, followed by a low rank matrix completion algorithm \cite{keshavan2009matrix}. Their algorithm \emph{produces a ranking}. On the other hand, our goal is mixture learning and the resulting theoretical analysis is different. 

\subsection{Iterative Refinement via EM}\label{subsect:em-refinement}

\textbf{The Weighted LSR Algorithm.} As noted before, we wish to maximize the weighted log-likelihood (\ref{eqn:weighted-log-likelihood}) efficiently. Towards this goal, we generalize the Luce spectral ranking (LSR) algorithm \citep{maystre2015fast} to incorporates sample weights. The original LSR algorithm produces the MLE. Our generalized algorithm outputs the weighted MLE (see Theorem \ref{thm:weighted-lsr}).

The intuition behind LSR is an interpretation of the \emph{PL ranking generative process as a sequence of choices} \cite{plackett1975analysis}.
Given a ranking $\pi_l$, define its \emph{choice breaking} as 
$$ \B_{\pi_l} = \{ (\pi_{l,1}, \{\pi_{l,1},  \ldots, \pi_{l,n}\}, l), \ldots , (\pi_{l,n-1}, \{\pi_{l,n-1}, \pi_{l,n}\}, l)   \} \,.$$
Each tuple $(i, A, l) \in \B(\pi_l)$ is a \emph{choice enumeration} of the ranking $\pi_l$. 
Given a ranking dataset $\Pi = \{\pi_1, \ldots, \pi_m\}$, define the \emph{choice breaking} of $\Pi$ as the union of all ranking-level choice breakings:
\begin{equation}\label{eqn:choice-breaking} \B_{\Pi} = \B_{\pi_1} \cup \ldots \cup \B_{\pi_m} \,.\end{equation}
Note that $\lvert \B_{\Pi}\rvert = m(n-1)$.
When the dataset $\Pi$ is clear from context, we simply use $\B$ to denote the dataset-level choice breaking. 

\emph{We now introduce sample weights}. Firstly, define the `weight' of a choice breaking $\B$ with weight vector $w$ and parameter $\theta \in \R^n$ as
\begin{equation}\label{eqn:gamma-weighted}
\gamma(\B, w, \theta) = w^\top \bigg(\frac{1}{\sum_{j\in A} e^{\theta_j} }\bigg)_{(i,A, l)\in \B} \,,
\end{equation}
where $w \in \R^{m(n-1)}_+$ is an arbitrary weight vector; $\bigg(\frac{1}{\sum_{j\in A} e^{\theta_j} }\bigg)_{(i,A,l)\in \B}$ is also vector in $ \R^{m(n-1)}_+$ where each entry corresponds to a choice enumeration $(i, A, l)$.

The reader may recognize that the weight vector $w$ has the same size as the choice breaking while sample weights are often given at the ranking level -- each ranking $\pi_l$ is assigned a weight $q_l$ for $l\in [m]$ as in (\ref{eqn:weighted-log-likelihood}). Given sample weights $q = (q_1, \ldots, q_m)$, one simply sets
\begin{equation}\label{eqn:w-from-q}
w = [ \underbrace{q_1, \ldots, q_1}_{n-1 \text{ terms}}, \underbrace{q_2, \ldots, q_2}_{n-1 \text{ terms}}, \ldots, \underbrace{q_m, \ldots, q_m}_{n-1 \text{ terms}}]^\top\,.
\end{equation}
Given a choice breaking $\B$ and items $i, j$, define the set of choice enumerations where $i$ `beats' $j$ as
$$ \B_{i\succ j} = \{ (i, A, l) \in \B \,:\, j \in A \}\,. $$
As a shorthand notation, for a weight vector $w$ corresponding to choice breaking $\B$, define $w_{j\succ i}$ as the sub-vector of $w$ corresponding to $\B_{i\succ j}$. 

Similarly to the original LSR algorithm, we construct a Markov chain (MC) and recover PL parameters from its stationary distribution. This MC has $n$ states. Given choice breaking $\B$, weight vector $w$ and parameter $\theta$, the pairwise transition probabilities of $M$ are given as
\begin{equation}\label{eqn:markov-chain}
    M_{ij} = \begin{cases}
    \frac{1}{d} \cdot \gamma (\B_{j \succ i},{w_{j\succ i} }, \theta) &\text{if } i\neq j\\
    1 - \frac{1}{d}\cdot \sum_{k\neq i} \gamma(\B_{k \succ i},w_{k\succ i}, \theta) &\text{if } i= j
    \end{cases}\,,
\end{equation}
where $d$ is a sufficiently large normalization constant such that $M$ does not contain any negative entries. Intuitively, $M_{ij}$ is proportional to the sum of the weights of all choice enumerations where $j$ `beats' $i$.

Algorithm \ref{alg:weighted-luce} summarizes the weighted LSR algorithm. It repeatedly constructs a Markov chain based on the current estimate, computes its stationary distribution and recovers the next estimate until convergence. When sample weights are not given, the weighted LSR algorithm reduces to the original LSR algorithm.

\begin{algorithm}[h]
\caption{Weighted Luce Spectral Ranking}
\hspace*{\algorithmicindent} \textbf{Input: } Dataset $\Pi = \{\pi_1, \ldots, \pi_m\}$, (optional) weight vector $q \in \R^{m}_{+}$ and (optional) initial estimate $\hat\theta^{(0)} \in \R^n$. \\
\hspace*{\algorithmicindent} \textbf{Output: } Normalized estimate of the item parameters $\hat\theta \in \R^n$.\\
\vspace{-0.75em}
\begin{algorithmic}[1]
    \STATE Obtain choice breaking $\B$ from $\Pi$ per Equation (\ref{eqn:choice-breaking}). 
    \STATE If the weight vector $q$ is not given, set $q = \mb{1}_m$.
    \STATE Construct $w$ from $q$ per Equation (\ref{eqn:w-from-q}).
    \STATE If the initial estimate is not given, set $\hat\theta^{(0)} =\mb{0}_n$.
    \STATE For $t = 1, \ldots$ until convergence\\
    \quad 5.1: Construct a Markov chain $M$ with pairwise transition probability per Equation (\ref{eqn:markov-chain}) from choice breaking $\B$, weight vector $w$ and parameter $\hat\theta^{(t-1)}$.\\
    \quad 5.2: Compute the stationary distribution of $M$ (e.g., via power iteration), $p$ and return the normalized estimate $\hat \theta^{(t)} = \log(p) - \big(\frac{1}{n} \sum_{i=1}^n \log(p)\big) \cdot \mb 1_n$ \\
\end{algorithmic}
\label{alg:weighted-luce}
\end{algorithm}

\textbf{The EM-LSR Algorithm.} In the E-step, we compute the posterior class probabilities $q^k \in \R^{m}, k\in [K]$. The M step consists of $K$ maximization problem as shown in Equation (\ref{eqn:weighted-log-likelihood}). These can be solved in parallel by running Algorithm \ref{alg:weighted-luce} on $\Pi$ using $q^k$ as sample weights for $k\in [K]$. Algorithm \ref{alg:spectral-em} summarizes the overall algorithm.
\begin{algorithm}[h]
\caption{Spectral EM (EM-LSR)}
\hspace*{\algorithmicindent} \textbf{Input: } Dataset $\Pi = \{\pi_1, \ldots, \pi_m\}$, number of components $K$, prior distribution $\beta$, (optional) initial estimate $\hat\btheta^{(0)} \in \R^{n\times K} $. \\
\hspace*{\algorithmicindent} \textbf{Output: } Normalized estimate $\hat\btheta = [\hat\theta^1, \ldots, \hat\theta^K]$.\\
\vspace{-0.75em}
\begin{algorithmic}[1]
    \STATE If $\hat\btheta^{(0)}$ is not given, run Algorithm \ref{alg:spectral-init} on $\Pi$ with $K$ mixture components and set $\hat\btheta^{(0)}$ to the output.
    \STATE For $t = 1,2,\ldots$ until convergence \\
    \quad 2.1: E-step -- Compute the class posterior probabilities $q^k_l = p(z^*_l = k|\pi_l, \hat\btheta^{(t-1)})$ for $l\in [m], k\in [K]$.\\
    \quad 2.2: M-step -- Estimate $\hat\theta^{k(t)}$ by running Algorithm \ref{alg:weighted-luce} on $\Pi$ with sample weight vector $q^k = [q^k_1,\ldots, q^k_m]$ and initial estimate $\hat\theta^{k(t-1)}$ for $k\in [K]$. \\
\end{algorithmic}
\label{alg:spectral-em}
\end{algorithm}

In another EM-based approach for learning PL mixtures, \citet{liu2019learning} use the \emph{unweighted LSR} algorithm. There, the E-step remains the same. The key differences lie in initialization (they use random initialization) and in the M-step. Our algorithm maximizes the weighted log-likelihood via weighted LSR and is therefore an exact EM algorithm. On the other hand, Liu et al. use the posterior class probabilities to perform a random clustering of the rankings and then run unweighted LSR on each cluster, making their algorithm an \emph{inexact} EM algorithm. From additional experiments in the supplementary materials, one can observe that the stochastic M-step actually leads to \emph{worse estimates without a significant reduction in inference time}.

\section{Theoretical Analysis}

In this section, we study the theoretical properties of EM-LSR. Section \ref{subsect:spectral-init} presents the finite sample error guarantee for the spectral initialization algorithm. Section \ref{subsect:em} focuses on the analysis of the M-step.

\subsection{Spectral Initialization}\label{subsect:spectral-init}
Central to the analysis of the spectral initialization algorithm is the accuracy of spectral clustering (Algorithm \ref{alg:spectral-clustering}). Our analysis starts from the fact that, under the pairwise representation in Equation (\ref{eqn:embedding}), the PL distribution exhibits \emph{sub-gaussian characteristics} \cite{vershynin2018high,shah2018learning}. The detailed descriptions of these characteristics are not immediately important to our discussions so we refer the interested reader to the supplementary materials. However, we emphasize that these characteristics also appear in a broad class of ranking models known as \emph{random utility models} (RUMs) that subsume the PL model. The spectral clustering algorithm is model-agnostic. It can be applied to mixtures of sub-gaussian distributions and enjoys high clustering accuracy if the signal-to-noise ratio (SNR) is high. We also show how, by changing the mapping function used in Algorithm \ref{alg:least-squares-estimate}, we can perform parameter estimation for a general RUM, not just PL. Thanks to this flexibility, Algorithm \ref{alg:spectral-init} can be a useful tool for learning mixtures of general RUMs.

% Algorithm \ref{alg:spectral-clustering} is therefore applicable to a much broader class of mixtures of ranking models. In the supplementary materials, we also show how, by changing the mapping function used in Algorithm \ref{alg:least-squares-estimate}, we can apply the spectral initialization algorithm to learn a mixture of general RUMs.
We now consider an expressive generative model for mixtures of $K$ PLs where Algorithm \ref{alg:spectral-init} produces a provably accurate estimate. The generative model assumes that for all mixture components, only the utilities of the first $L$ items are different while the those of the remaining $n-L$ items are the same. This model reflects the phenomenon where users from different sub-populations differ in their preference among a few items while the remaining items are essentially interchangeable. Intuitively, one would expect that when $L$ is small, so is the difference between the subpopulations and it is harder to separate the rankings into the correct clusters. On the other hand, when $L$ is large, the difference among the subpopulations is large and it is easier to separate the clusters. The following theorem captures this intuition.

\begin{theorem}\label{thm:top-L-mixture-error-init} Consider a mixture of $K$ Plackett-Luce models with uniform mixing probabilities. Suppose that $\theta^{*k}_{i} = 0 \,\forall i \in [L+1:n]$ and $\theta^{*k}_{1:L} \sim \N(0, I_L)$ for $k\in [K]$. Fix a constant $\alpha > 0$. There exist constants $c, c_1, C_1, C_2, D$ such that if $m\geq c\max \{K^4, Kn\}$ then the output $\hat\btheta$ of Algorithm \ref{alg:spectral-init} satisfies the following. 
If $L \geq c_1\,\exp{C_1 \sqrt{\log n}}$, then
\begin{multline*} 
\hspace{-0.5cm}\text{dist}(\hat\btheta, \btheta^*) = O\bigg(\exp{D\sqrt{\log n}} \bigg(\sqrt{\frac{K^2n\log n}{m}} + \frac{\sqrt{Kn}}{e^{ L^{0.99}}} \bigg)  \bigg)\,
\end{multline*}
with probability $1- O(\frac{K}{n^8}) - O(K^2n^2\exp{-L^{0.99}})$. 
If $L \geq C_2 n^\alpha$ and assuming that $n = \omega(\log m)$, then
\begin{equation*}
    \text{dist}(\hat\btheta, \btheta^*) = O\bigg( \exp{D\sqrt{\log n}}\, \sqrt{\frac{K^2n\log n}{m}}\, \bigg)\,
\end{equation*}
with probability $1- O(\frac{K}{n^8}) - O(K^2n^2\exp{-n^{\alpha}})$.
\end{theorem}
The first error bound is a sum of two terms. The first is the estimation error incurred by Algorithm \ref{alg:least-squares-estimate} which diminishes with increasing $m$. The second comes from the clustering error incurred by Algorithm \ref{alg:spectral-clustering} and is controlled by the SNR of the generative model. One can also check that $\exp{\sqrt{\log n}} = o(n^\alpha)$ for any $\alpha > 0$ and $\exp{\sqrt{\log n}} = \omega(\log n)$.
When $L \approx \exp{O(\sqrt{\log n})}$ (low SNR), there is significant clustering error and the second term scales approximately as $O(\frac{\sqrt{n}}{e^{L}}) = O\big(\frac{1}{\poly(n)}\big)$.
Hence, Algorithm \ref{alg:spectral-init} converges to within a small radius around $\btheta^*$ given a sufficiently large $m$. However, when $L$ is polynomial in $n$ (high SNR), estimation error dominates clustering error, giving us the second error bound which diminishes with sample size $m$. In this regime, the spectral initialization algorithm works well as a \emph{standalone mixture learning algorithm}. Note that this guarantee holds even for a small $\alpha > 0$, when the fraction of `informative' items diminishes: $L/n = o(1)$. Our proposed generative model is new and could be a useful analysis framework for future works. To the best of our knowledge, the finite sample error bounds are also the \emph{first of their kind in the literature}.

\subsection{Iterative Refinement via EM}\label{subsect:em}
\textbf{Accuracy of the M-step.} The following theorem generalizes Theorem 1 of \citet{maystre2015fast}.
% The following theorem establishes that the estimate given by Algorithm \ref{alg:weighted-luce} is the maximizer of the \emph{weighted log-likelihood function}. Our theorem generalizes Theorem 1 of \citet{maystre2015fast}.
\begin{theorem}\label{thm:weighted-lsr} The output of weighted LSR (Algorithm \ref{alg:weighted-luce}) is the maximum weighted log-likelihood estimate:
$$ \theta_q^{\text{MLE}} := \arg\max_{\theta} \sum_{l=1}^m \, q_l \cdot \log \Pr^{\text{PL}}(\pi_l, z_l \,\lvert\, \theta ) \,. $$
% where $ \Pr^{PL}(\pi_l\,\lvert \theta )$ is shown in Equation (\ref{eqn:plackett-luce-likelihood}).
\end{theorem}
% In the context of the EM-algorithm, each M-step involves solving $K$ separate maximization problems where the objective functions are weighted log-likelihood functions with weights given by the class posterior distributions $q$.
As noted before, the EMM algorithm is an alternative approach that exactly solves the M-step using the (weighted) MM algorithm. 
In other words, \emph{assuming perfect numerical precision and the same initialization}, EMM and EM-LSR will produce the same final estimate. However, our EM-LSR algorithm is often much faster than EMM (e.g., Figure \ref{fig:synthetic}).

\textbf{Convergence of EM.} It is well known that the EM algorithm converges to a stationary point \cite{wu1983convergence}. There is, unfortunately, no guarantee how close such a point is to the global optimum. However, assuming correct model specification and that the initial estimate falls within a neighborhood around $\theta^*$ which satisfies certain high SNR conditions, the EM algorithm will converge to $\theta^*$ \cite{wang2015high,wu2016convergence,balakrishnan2017statistical}. The area around $\theta^*$ where this desirable behaviour occurs is referred to as the \emph{basin of attraction}.  We leave the detailed characterization of the basin of attraction as a subject of future studies.

\textbf{True Likelihood versus Surrogate Likelihood.} For two other commonly used EM algorithms in the literature -- EM-CML and EM-GMM -- previous authors use random initialization. On the other hand, ours uses spectral initialization. However, initialization is not the only differentiating characteristic of our algorithm. In fact, our algorithm, EM-CML and EM-GMM are \emph{fundamentally different EM-based algorithms}. To see why, one needs to inspect the objective function of the M-step. Suppose that all three algorithms are initialized at some $\hat\btheta^{(0)}$. Let $\{q^k_{l} \}_{l\in[m]}^{k\in [K]}$ denote the posterior class probabilities conditioned on $\hat\btheta^{(0)}$ per Equation (\ref{eqn:posterior}).

In the first iteration, EM-LSR and EMM maximize the weighted log-likelihood.
\begin{equation*}
\hat\btheta_{\lsr}^{(1)} =\arg\max_{\btheta} \sum_{l=1}^m \sum_{k=1}^K \bigg[q^k_{l}\cdot \log \Pr^{PL}(\pi_l, z_l\,\lvert\, \theta^k )  \bigg] \,.
\end{equation*}
On the other hand, EM-CML maximizes the \emph{composite (surrogate) marginal likelihood}. $\hat\btheta_{\cml}^{(1)} = \arg\max_{\btheta}$
\begin{multline*}
\sum_{l=1}^m \sum_{k=1}^K \bigg[\sum_{\substack{i,j:\\\pi_l(i) < \pi_l(j)}} q^k_{l} \log\bigg(\frac{1}{1+\exp{-(\theta^k_i -\theta^k_j )}}  \bigg)   \bigg]\,.
\end{multline*}
Lastly, EM-GMM \emph{minimizes} the following function.
\begin{equation*}
\hat\btheta_{\gmm}^{(1)} = \arg\min_{\btheta} \sum_{k=1}^K \sum_{i \neq j} \bigg( \hat F^k_{ij} - \frac{1}{1 + \exp{-(\theta^k_i - \theta^k_j)}} \bigg)^2 \,,
\end{equation*}
where $\hat F^k_{ij} = \frac{\sum_{l=1}^m \mb 1[\pi_l(i) < \pi_l(j)] \,q_l^k }{\sum_{l=1}^m q_l^k }$.

One can see that the objective functions are different and so are their solutions. Hence, even if we initialize all three algorithms with the same estimate, their trajectories will be different in general. While EM-LSR and EMM converges to the true MLE when initialized within the basin of attraction, this \emph{may not be true for EM-GMM and EM-CML}. This difference is supported by our experiments, where even with the same initialization, the algorithms produce different final estimates.

\section{Experiments}
We compare our spectral EM algorithm to the following baselines: EMM, EM-GMM and EM-CML.

\begin{figure}[]
    \centering
    \begin{subfigure}[h]{0.30\textwidth}
        \centering
        \includegraphics[scale=0.4]{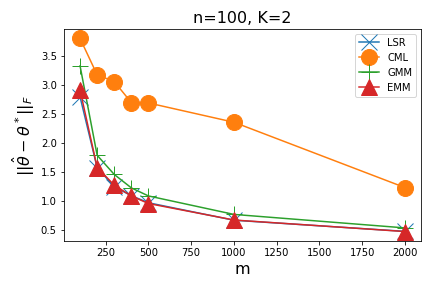}
        \caption{For a small number of mixture components, all methods are quite accurate. As the theory implies, EMM and EM-LSR produce similar estimates. \label{fig:syn-small-error}}
    \end{subfigure}%

    \begin{subfigure}[h]{0.30\textwidth}
        \centering
        \includegraphics[scale=0.4]{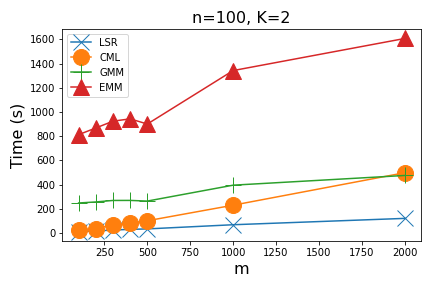}
        \caption{EM-LSR is comparatively efficient (figure shows total inference time). \label{fig:syn-small-time}}
    \end{subfigure}%
    
    \begin{subfigure}[h]{0.30\textwidth}
        \centering
        \includegraphics[scale=0.4]{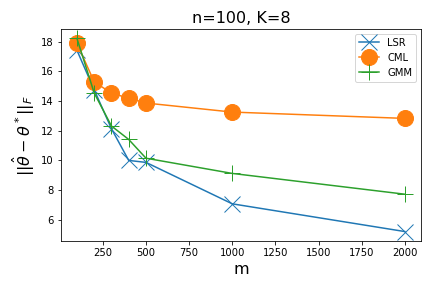}
        \caption{For a moderate number of mixture components, EM-LSR is the most accurate method (EMM not shown due to timeout).\label{fig:syn-mod-error}}
    \end{subfigure}

    \begin{subfigure}[h]{0.30\textwidth}
        \centering
        \includegraphics[scale=0.4]{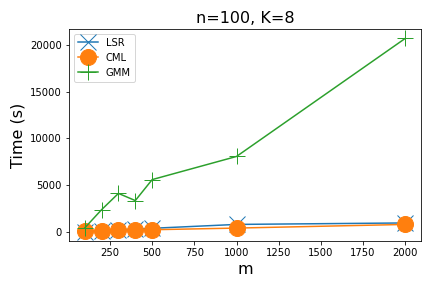}
        \caption{EM-LSR and EM-CML are the only two methods that are efficient for a moderate number of mixture components. \label{fig:syn-mod-time}}
    \end{subfigure}
    \caption{$\ell_2$ error and inference time on \textbf{synthetic datasets}. EM-LSR (in \textcolor{blue}{blue}) is competitive in terms of accuracy and speed to the baseline algorithms. \label{fig:synthetic}}
\end{figure}

\textbf{Synthetic Datasets.} We simulate data from the generative model as described in Theorem \ref{thm:top-L-mixture-error-init}. Specifically, we set $n=100$ and $L=5$ while varying the number of mixture components $K$ for different experiments. Figure \ref{fig:synthetic} shows estimation error and total inference time against the sample size $m$, averaged over 25 trials. Experimentally, spectral initialization consistently gives better initial estimates than both random initialization and GMM initialization \cite{zhao2016learning}. To keep a fair comparison, we use spectral initialization for all algorithms. When $K$ is small (e.g., Figures \ref{fig:syn-small-error} and \ref{fig:syn-small-time}) all four methods are quite accurate. When the number of mixture components are moderate (e.g., Figures \ref{fig:syn-mod-error} and \ref{fig:syn-mod-time}), the advantages that EM-LSR enjoys over the other methods become more apparent. While EMM becomes too inefficient for practical purposes, EM-LSR remains relatively efficient and produces more accurate estimates than both EM-CML and EM-GMM. 

\textbf{Real Datasets.} We include commonly used datasets in previous works such as APA, Irish Elections (West, North, Meath) and SUSHI all with $n < 15$. We partition all the rankings with a 80-20 training-testing split; and the train rankings into 80\% for inference and 20\% for validation. $K$ is chosen using Bayesian Information Criterion \cite{gelman2014understanding} on the validation set and the log-likelihood of the final model is evaluated using the test set. For these datasets, EM-LSR and EMM are the most accurate while EM-CML is the fastest, especially on datasets with a large $m$ such as the Irish election datasets. We have a possible explanation for the relative speed between EM-LSR and EM-CML. The bottle neck in these EM algorithms is the M-step. The most time-consuming procedure in the M-step of EM-LSR is constructing the Markov chain in Algorithm \ref{alg:weighted-luce} with time complexity $O(mn^2)$. For EM-CML, it is solving a constrained concave maximization problem via SLSQP \cite{scipy} and may scale at least as $\Omega(n^3)$\footnote{SLSQP solves a sequence of quadratic optimization problems with $n$ variables. Each solves a linear system with $n$ variables and $n$ equations and generally takes $O(n^3)$ \cite{strang1993introduction}.}. Therefore, EM-CML tends to be faster for datasets with a small $n$ and a large $m$. However, its inference time could grow significantly with $n$.

\begin{figure}[]
    \centering
    \begin{subfigure}[h]{0.30\textwidth}
        \centering
        \includegraphics[scale=0.4]{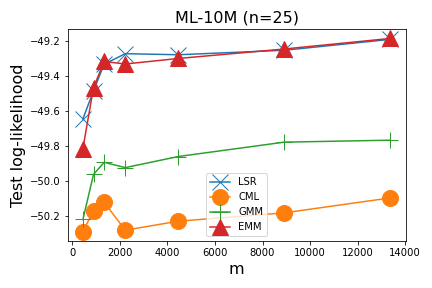}
        \caption{With a small number of items, EMM and EM-LSR are more accurate.}
    \end{subfigure}%

    \begin{subfigure}[h]{0.30\textwidth}
        \centering
        \includegraphics[scale=0.4]{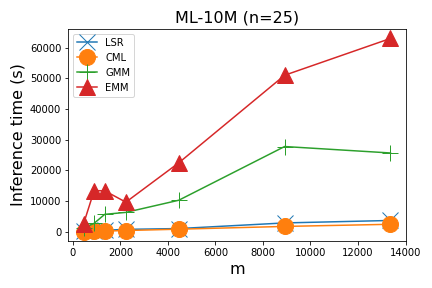}
        \caption{However, EM-LSR is also competitive in terms of efficiency.}
    \end{subfigure}

    \begin{subfigure}[h]{0.30\textwidth}
        \centering
        \includegraphics[scale=0.4]{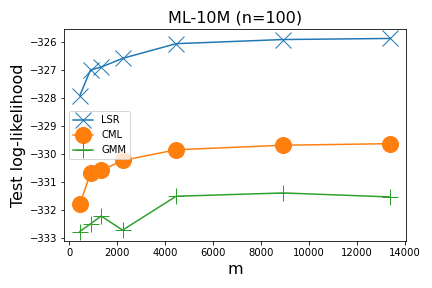}
        \caption{For a larger set of items, EM-LSR is the most accurate method (EMM not shown due to timeout).}
    \end{subfigure}

    \begin{subfigure}[h]{0.30\textwidth}
        \centering
        \includegraphics[scale=0.4]{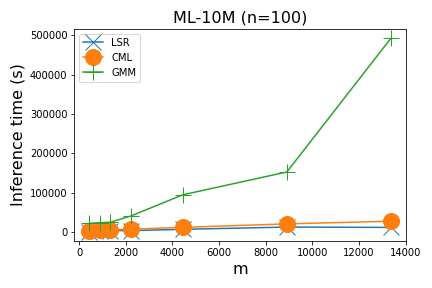}
        \caption{EM-LSR is comparatively scalable for larger datasets.}
    \end{subfigure}

    \caption{Test log-likelihood and inference time on \textbf{ML-10M datasets}. For larger datasets, EM-LSR (in \textcolor{blue}{blue}) is more accurate while being competitive in speed to the baseline algorithms. \label{fig:ml-10M}}
\end{figure}

Indeed, the setting where EM-LSR outperforms the baselines is when $n$ is large. We perform additional experiments on the ML-10M movie ratings datasets \cite{harper2015movielens}. To generate rankings, we first run a low rank matrix completion algorithm \cite{Zitnik2012} on the user-item rating matrix to fill in the missing entries. We then select $n$ movies from the set of all movies and the rankings are obtained from the completed matrix. Figure \ref{fig:ml-10M} shows the performance of the four methods on two versions of the ML-10M datasets with $n=25$ and $n=100$ given increasing training data up to 14k. In the supplementary materials, we also include additional experiments, strategies to extend EM-LSR to handle partial rankings with ties and comparisons to a Bayesian method \cite{mollica2017bayesian}.

\begin{table}[h]
\hspace{-0.5cm}
\resizebox{1.15\columnwidth}{!}{%
\begin{tabular}{|c|cccc|cccc|}
\hline
Dataset & \multicolumn{4}{c|}{Test log-likelihood}                                             & \multicolumn{4}{c|}{Inference time (s)}                                            \\ \hline
        & \multicolumn{1}{c|}{EM-LSR} & \multicolumn{1}{c|}{EM-CML} & \multicolumn{1}{c|}{EM-GMM} & EMM & \multicolumn{1}{c|}{EM-LSR} & \multicolumn{1}{c|}{EM-CML} & \multicolumn{1}{c|}{EM-GMM} & EMM \\ \hline
APA ($n=5$) & \multicolumn{1}{c|}{-4.619}  & \multicolumn{1}{c|}{-4.656} & \multicolumn{1}{c|}{-4.617} & \shade -4.614 & \multicolumn{1}{c|}{598}  & \multicolumn{1}{c|}{\shade 33.3} & \multicolumn{1}{c|}{2.2\text{K}} & 9.24\text{K} \\ \hline
West ($n=9$)  & \multicolumn{1}{c|}{-11.9}  & \multicolumn{1}{c|}{-12.008} & \multicolumn{1}{c|}{-11.904} & \shade -11.896 & \multicolumn{1}{c|}{810}  & \multicolumn{1}{c|}{\shade 199} & \multicolumn{1}{c|}{5.75\text{K}} & 25.8\text{K} \\ \hline
Sushi ($n=10$) & \multicolumn{1}{c|}{\shade-13.64}  & \multicolumn{1}{c|}{-14.0} & \multicolumn{1}{c|}{-13.773} & -13.766 & \multicolumn{1}{c|}{746}  & \multicolumn{1}{c|}{\shade 24.6} & \multicolumn{1}{c|}{489} & 1.22\text{K} \\ \hline
North ($n=12$) & \multicolumn{1}{c|}{\shade-18.67}  & \multicolumn{1}{c|}{-18.923} & \multicolumn{1}{c|}{-18.744} & -18.711 & \multicolumn{1}{c|}{1.51\text{K}}  & \multicolumn{1}{c|}{ \shade 120} & \multicolumn{1}{c|}{3.09\text{K}} & 14\text{K} \\ \hline
Meath ($n=14$) & \multicolumn{1}{c|}{-23.645}  & \multicolumn{1}{c|}{-23.885} & \multicolumn{1}{c|}{-23.69} & \shade-23.633 & \multicolumn{1}{c|}{1.48\text{K}}  & \multicolumn{1}{c|}{\shade 497} & \multicolumn{1}{c|}{29.9\text{K}} & 69.1\text{K} \\ \hline
\hline
ML-10M ($n=25$) & \multicolumn{1}{c|}{-49.191}  & \multicolumn{1}{c|}{-50.095} & \multicolumn{1}{c|}{-49.766} & \shade -49.186 & \multicolumn{1}{c|}{3.71\text{K}}  & \multicolumn{1}{c|}{\shade 2.47\text{K}} & \multicolumn{1}{c|}{25.7\text{K}} & 63\text{K} \\ \hline
ML-10M ($n=50$) & \multicolumn{1}{c|}{\shade -130.499}  & \multicolumn{1}{c|}{-132.209} & \multicolumn{1}{c|}{-132.143} & NA & \multicolumn{1}{c|}{\shade 5.77\text{K}}  & \multicolumn{1}{c|}{6.8\text{K}} & \multicolumn{1}{c|}{125\text{K}} & NA \\ \hline
ML-10M ($n=100$) & \multicolumn{1}{c|}{\shade -325.873}  & \multicolumn{1}{c|}{-329.629} & \multicolumn{1}{c|}{-331.53} & NA & \multicolumn{1}{c|}{\shade 11.5\text{K}}  & \multicolumn{1}{c|}{27.2\text{K}} & \multicolumn{1}{c|}{492\text{K}} & NA \\ \hline
ML-10M ($n=150$) & \multicolumn{1}{c|}{\shade -550.462}  & \multicolumn{1}{c|}{-557.923} & \multicolumn{1}{c|}{NA} & NA & \multicolumn{1}{c|}{\shade 14.7\text{K}}  & \multicolumn{1}{c|}{62.1\text{K}} & \multicolumn{1}{c|}{NA} & NA \\ \hline
ML-10M ($n=200$)  & \multicolumn{1}{c|}{\shade -787.244}  & \multicolumn{1}{c|}{-799.036} & \multicolumn{1}{c|}{NA} & NA & \multicolumn{1}{c|}{\shade 24.5\text{K}}  & \multicolumn{1}{c|}{81.3\text{K}} & \multicolumn{1}{c|}{NA} & NA \\ \hline
\end{tabular}
}
\caption{Test log-likelihood and inference time on real datasets. `NA' denotes not available due to timeout. \label{tbl:small-datasets}}
\end{table}

% As seen in Figure \ref{fig:ml-10M} and the second half of Table \ref{tbl:small-datasets}, for datasets with at least a moderate number of items $n \geq 25$, EM-LSR is often the best performing method in terms of both accuracy and efficiency.

\vspace{-0.25cm}
\section{Conclusion}
We have proposed an accurate and efficient algorithm for learning a mixture of Plackett-Luce models. For future works, we would like to consider other initialization methods such as the method of moments or tensor decomposition.
Detailed characterization of the basin of attraction within which the EM algorithm converges to the true parameter is also a challenging open question. On a more practical note, incorporating the representation power of deep neural networks into our algorithm will further increase its utility for large scale recommendation systems applications.

\section{Acknowledgements}

The authors thank the anonymous reviewers for their thoughtful suggestions and comments. The authors are supported by NSF Grant DMS-2112099. A.Z. acknowledges financial support from the Alfred H. Williams Faculty Scholar award. Any opinions expressed in this paper are those of the authors and do not necessarily reflect the views of the National Science Foundation.

\bibliography{references}

% \newpage
% \appendix
% \onecolumn
% % \addcontentsline{toc}{section}{Appendix} % Add the appendix text to the document TOC
% % \part{Appendix} % Start the appendix part
% % \parttoc % Insert the appendix TOC

% \appendixpage

% \startcontents[chapter]
% \printcontents[chapter]{l}{0}{\setcounter{tocdepth}{1}}

% \import{./proofs/}{proofs}

% % \newpage 
% \bibliographystyle{plainnat}
% \bibliography{references}
% % \bibliographystyle{abbrv}

\end{document}

\makeatletter\@input{extra.tex}\makeatother

\maketitle
\appendix

\tableofcontents

\section{Preliminaries} 
In this section, we introduce extra definitions and notations that will be used throughout the rest of the paper and some useful high-dimension concentration inequalities. Many of these results apply not only to the Plackett-Luce model but also other \emph{random utility models} (see Definition \ref{def:rum}). We believe these results can be of independent interest.

\subsection{Useful Concentration Inequalities}
As a shorthand notation, $a \lesssim b$ means that $a \leq C b$ for some constant $C$ and $a \asymp b$ means that $a\lesssim b, b \lesssim a$.

\begin{definition}[Random utility model generative process]\label{def:rum} A random utility model of ranking data is parametrized by utilities $\theta^*= [\theta^*_1, \ldots, \theta^*_n]$ and noise distributions $\calD_1, \ldots, \calD_n$. A ranking $\pi$ is drawn by first drawing $\epsilon_i \sim \calD_i$ independently. Define $U_i = \theta^*_i + \epsilon_i$ for $i \in [n]$. $\pi = \arg\text{\emph{sort}} \, U$.
\end{definition}

When all the noise distributions are the same $\calD_i = \calD$ for some distribution $\calD$, the ranking distribution is also referred to as a random utility model with independently and identically distributed noise, or IID-RUM. The PL model is an IID-RUM with a standard Gumbel noise distribution. The Thurstone model \citep{thurstone2017law} is 
another IID-RUM but with $\calN(0,\frac{1}{2})$ being the noise distribution.

\begin{definition}[Sub-gaussian random variable, Proposition 2.5.2 of \cite{vershynin2018high}] A random variable $X \in \R$ is a sub-gaussian random variable with variance parameter $\sigma$ if the tails of $X$ satisfies
$$ \Pr\bigg( \lvert X - \E X \rvert \geq t  \bigg) \leq 2\, \exp{-\frac{t^2}{\sigma^2}} $$
and the sub-gaussian norm of $X$ is defined as
$$ \lVert X\rVert_{\psi_2} = \inf\{t > 0\,:\, \E\,\exp{(X-\E X)^2/t^2} \leq 2 \}. $$
\end{definition}

\begin{definition}[Sub-gaussian distribution in high dimension, Definition 3.4.1 of \cite{vershynin2018high}] A random vector $X \in \R^d$ is a sub-gaussian random variable if the one-dimensional marginals $X^\top x$ are sub-gaussian random variables for all $x\in \R^d$. The sub-gaussian norm of $X$ is defined as
$$ \lVert X\rVert_{\psi_2} = \sup_{x: \lVert x \rVert_2 = 1} \,\lVert X^\top x\rVert_{\psi_2}\,. $$
\end{definition}

If a random vector $X$ is a sub-gaussian random vector with sub-gaussian norm $\sigma$, we write $X \sim \SG(\sigma^2)$.

\begin{lemma}[Sub-gaussian norm of permutation vector, Proposition 3 of \cite{shah2018learning}]\label{lem:sub-gaussian-norm-general} Consider the pairwise vector representation of permutations described in Equation (\ref{eqn:embedding}). Under any random utility model, there exists a constant $\tau$ such that the sub-gaussian norm of the random variable $X$ from such embedding satisfies
$$ \lVert X \rVert_{\psi} \leq \sqrt{\tau n} \,.$$
\end{lemma}

\begin{lemma}(Concentration of norm for sub-gaussian random variables, Theorem 3.1.1 of \cite{vershynin2018high})\label{lem:concentrate-norm} Let $X = (X_1, \ldots X_n)\in \R^n \sim \calN(0, \sigma_0 I_n)$. Then there exists a constant $c$ such that
$$ \Pr\bigg( \big\lvert \lVert X \rVert_2 - \sqrt n \big\rvert \geq t \bigg) \,\leq \,2\exp{-\frac{ct^2}{\sigma_0^4}} \,.$$
\end{lemma}

\begin{theorem}(Hoeffding's inequality) Let $X_1, X_2, \ldots, X_m$ be independent random variables such that $X_i \in [a_i, b_i]$ for $i \in [m]$. Then
$$ \Pr\bigg(\big\lvert \sum_{i=1}^m X_i - \E[X_i] \big\rvert > t\bigg) \leq 2\,\exp{-\frac{2t^2}{\sum_{i=1}^m (a_i - b_i)^2}} \,. $$
\end{theorem}

\begin{theorem}(Concentration inequality for a sum of Normal random variables)\label{thm:concentrate-normal-sum} Let $X_1, \ldots, X_n$ be independent normal random variables distributed as $\N(0, \sigma^2)$. Then
$$ \Pr\bigg(\big\lvert \sum_{i=1}^n X_i \big\rvert > t\bigg) \leq 2\exp{-\frac{t^2}{2\sigma^2 n}} \,. $$
\end{theorem}
% \begin{theorem}[Hanson-Wright's inequality \cite{rudelson2013hanson}]\label{thm:hanson-wright} Let $X$ be a $n$-dimensional random vector distributed according to $\calN(0, \sigma_0^2 I_n)$. Let $A$ be a $\R^{n\times n}$ diagonal free matrix (i.e., $A_{ii} = 0 \,\forall i \in [n]$). Then there exists a constant $c > 0$ such that
% $$ \Pr( \lvert X^\top A X \rvert \,>  t) \leq 2\exp{-c\min \bigg\{ \frac{t^2}{\sigma_0^4 \lVert A \rVert_F^2}, \frac{t}{\sigma_0^2 \lVert A \rVert_{op}} \bigg\}}\,. $$
% \end{theorem}

\subsection{Mixtures of Sub-gaussian Distributions}
We first introduce notations and theoretical quantities that bridge the gap between the analysis of mixtures of PL models and that of mixtures of sub-gaussian distributions. Let the means of $K$ sub-gaussian distributions be $\mu^*_1, \ldots, \mu^*_K$ ($\mu^*_k \in \R^{{n\choose 2}}$ in our algorithm). The observed pairwise vector $X_l$ for $l \in [m]$ can be written as
$$ X_l = \mu^*_{z_l} + \epsilon_l $$
where $\epsilon_l \in \R^{{n\choose 2}} $ is a random noise vector with gaussian norm $\lesssim \sqrt n$ as shown earlier. Let $E = [\epsilon_1, \ldots, \epsilon_m] \in \R^{m\times {n\choose 2}} $ be the concatenation of all the noise vectors. 
Let $P \in \R^{m\times {n\choose 2}}$ be the concatenation of the `centers' of the $m$ random vectors: $P = [\E[X_1], \ldots, \E[X_m]]^\top $.
In short,
$$ X = P + E $$
Let $\Delta = \min_{k\neq k'} \lVert \mu^*_k - \mu^*_{k'}\rVert_2 $ denote the minimum inter-center gap.

~\\
The following lemma will be useful in deriving a bound on the operator norm of $E$. This lemma is often found in the analysis of random graphs such as in stochastic block model.
\begin{lemma}\label{lem:spectral-norm-bound-lei} [Theorem 5.2 of \cite{lei2015consistency}] Let $A$ be the adjacency matrix of a random graph on $n^*$ node in which edges occur independently. Set $P = \E[A]= (p)_{ij}$ and assume that $\max_{ij} p_{ij} \leq \frac{d}{n^*}$ for some $d \geq c_0\log n$ and $c_0 > 0$. Then for any $r > 0$, there is a constant $C$ that depends only on $r, c_0$ such that 
$$ \lVert E \rVert_{op} = \lVert A - P\rVert_{op}\leq C\sqrt d $$
with probability at least $1-{n^*}^{-r}$.
\end{lemma}

\begin{lemma}\label{lem:operator-norm-bound} Under the generative model for a mixture of $K$ random utility models, there exists a constant $C_E$ such that the operator norm of the error matrix satisfies
$$\lVert E \rVert_{op} \leq C_E \sqrt n (\sqrt{m+n}) \, $$
with probability at least $1-(m+n)^{-10}$.
\end{lemma}

\begin{proof} Note that in the broad class of random utility models, the distribution of the comparisons of two \emph{disjoint} pairs are independent. Building on this idea, we partition $E^\top$ into $O(n)$ blocks of size $O(n)\times m$ where each block consists of entries corresponding to pairs that are mutually disjoint. This is so that within each block, the entries (comparisons) are independent. Specifically, consider the set of all pairs 
$$\{ (1,2), (1,3), \ldots , (1,n), (2,3), \ldots, (2,n), \ldots, (n, n-1)  \} \,.$$
Imagine that we place all these pairs into the upper diagonal entries of a $n\times n$ matrix as shown in Figure \ref{fig:arranging-pairs}.
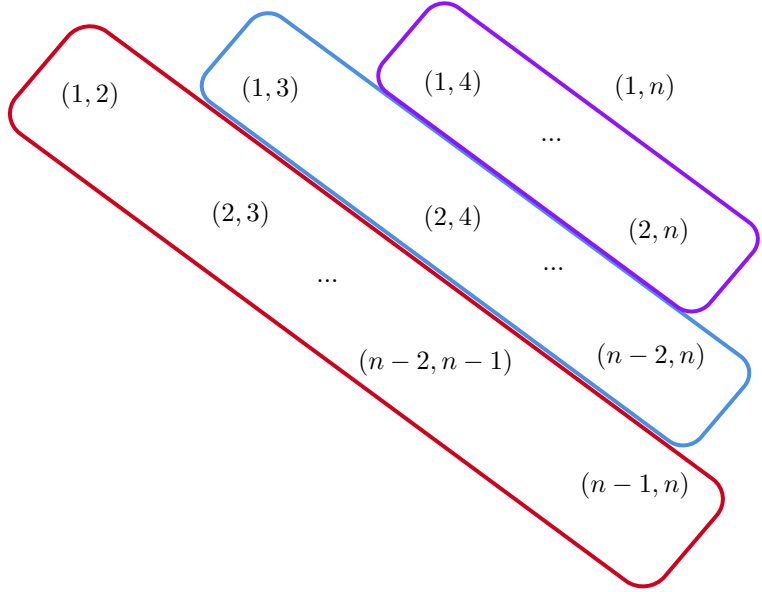
\begin{figure}[h]
\centering
\tikzset{every picture/.style={line width=0.75pt}} %set default line width to 0.75pt        

\begin{tikzpicture}[x=0.75pt,y=0.75pt,yscale=-1,xscale=1]
%uncomment if require: \path (0,431); %set diagram left start at 0, and has height of 431

%Rounded Rect [id:dp46352999229496494] 
\draw  [color={rgb, 255:red, 208; green, 2; blue, 27 }  ,draw opacity=1 ][line width=1.5]  (161.25,99.93) .. controls (166.08,94.25) and (174.83,93.22) .. (180.81,97.64) -- (486.21,323.29) .. controls (492.19,327.71) and (493.13,335.9) .. (488.3,341.58) -- (462.11,372.43) .. controls (457.29,378.11) and (448.53,379.13) .. (442.55,374.72) -- (137.15,149.07) .. controls (131.18,144.65) and (130.24,136.46) .. (135.06,130.78) -- cycle ;
%Rounded Rect [id:dp01575539104299839] 
\draw  [color={rgb, 255:red, 74; green, 144; blue, 226 }  ,draw opacity=1 ][line width=1.5]  (252.94,92.47) .. controls (256.94,87.76) and (264.2,86.91) .. (269.15,90.58) -- (500.4,261.44) .. controls (505.35,265.1) and (506.13,271.89) .. (502.13,276.6) -- (480.42,302.18) .. controls (476.42,306.89) and (469.16,307.74) .. (464.2,304.07) -- (232.96,133.21) .. controls (228,129.55) and (227.22,122.76) .. (231.22,118.05) -- cycle ;
%Rounded Rect [id:dp5270601752726083] 
\draw  [color={rgb, 255:red, 144; green, 19; blue, 254 }  ,draw opacity=1 ][line width=1.5]  (341.94,87.47) .. controls (345.94,82.76) and (353.2,81.91) .. (358.15,85.58) -- (504.72,193.87) .. controls (509.68,197.54) and (510.46,204.32) .. (506.46,209.03) -- (484.74,234.62) .. controls (480.74,239.33) and (473.48,240.17) .. (468.53,236.51) -- (321.96,128.21) .. controls (317,124.55) and (316.22,117.76) .. (320.22,113.05) -- cycle ;

% Text Node
\draw (156,121.4) node [anchor=north west][inner sep=0.75pt]    {$( 1,2)$};
% Text Node
\draw (232,181.4) node [anchor=north west][inner sep=0.75pt]    {$( 2,3)$};
% Text Node
\draw (435,117.4) node [anchor=north west][inner sep=0.75pt]    {$( 1,n)$};
% Text Node
\draw (339,115.4) node [anchor=north west][inner sep=0.75pt]    {$( 1,4)$};
% Text Node
\draw (247,118.4) node [anchor=north west][inner sep=0.75pt]    {$( 1,3)$};
% Text Node
\draw (339,183.4) node [anchor=north west][inner sep=0.75pt]    {$( 2,4)$};
% Text Node
\draw (442,190.4) node [anchor=north west][inner sep=0.75pt]    {$( 2,n)$};
% Text Node
\draw (306,256.4) node [anchor=north west][inner sep=0.75pt]    {$( n-2,n-1)$};
% Text Node
\draw (426,253.4) node [anchor=north west][inner sep=0.75pt]    {$( n-2,n)$};
% Text Node
\draw (418,318.4) node [anchor=north west][inner sep=0.75pt]    {$( n-1,n)$};
% Text Node
\draw (285,221) node [anchor=north west][inner sep=0.75pt]   [align=left] {...};
% Text Node
\draw (399,215) node [anchor=north west][inner sep=0.75pt]   [align=left] {...};
% Text Node
\draw (398,150) node [anchor=north west][inner sep=0.75pt]   [align=left] {...};

\end{tikzpicture}
\caption{${n\choose 2}$ pairs are arranged into the upper diagonal entries of a square matrix with the super-diagonals highlighted. \label{fig:arranging-pairs}}
\end{figure}

Now for each super-diagonal of this matrix, partition the entries into half by taking every other item. For example, for the first super-diagonal, partition the entries as follows (WLOG assuming that $n$ is even) to obtain two blocks of entries.
$$ \{(1,2), (3,4), \ldots, (n-1, n)\} \, \cup \, \{(2,3), (4,5), \ldots, (n-2, n-1)\} \,.$$
Repeat the same procedure for all super-diagonals and we obtain $2(n-1)-1$ of these sets. Note that these subsets are partitions of the pairs (columns of $E$), we can rewrite $E$ as a concatenation of block matrices, each corresponding to a subset of pairs.
$$ E^\top = \begin{bmatrix}E_1^\top\\ E_2^\top\\ \vdots \\ E_{O(n)}^\top  \end{bmatrix} \,.$$
Each submatrix $E_i$ is of dimension $O(n)\times m$. Note that permutation of the columns of $E$ does not change the operator norm. It follows that:
$$ \lVert E \rVert_{op} = \lVert E^\top \rVert_{op} \leq \sqrt{2n} \, \max_{i} \lVert E_i\rVert_{op}\,. $$

To bound $\lVert E_i\rVert_{op}$, we need to use Lemma \ref{lem:spectral-norm-bound-lei} with $d=n$. Using the symmetric dilation trick,
$$ \lVert E_i \rVert_{op} = \lVert \calS(E_i) \rVert_{op} = \bigg\lVert\begin{bmatrix} 0 &E_i\\ E_i^\top &0\end{bmatrix} \bigg\rVert_{op} \,.$$
The matrix $\calS(E_i)$ is a symmetric, binary matrix of size $O((n+m)) \times O((n+m))$ with independently distributed entries (like an adjacency matrix). By Lemma \ref{lem:spectral-norm-bound-lei}, there exists a constant $C_E$ such that, with probability at least $1-(n+m)^{-10}$,
$$ \lVert E_i \rVert_{op} = \lVert \calS(E_i)\rVert_{op} \leq 1/\sqrt{2} \,C_E \sqrt{n+m} \,.$$
This completes the proof.
\end{proof}

The following theorem provides guarantees on the performance of the spectral clustering algorithm with adaptive dimension reduction (Algorithm \ref{alg:spectral-clustering}) when applied to a mixture of $K$ random utility models. Before showing the theorem, let us introduce some definitions. Let $\xi = \frac{1}{m/K} \,\min_{k\in [K]} \,\lvert \{l\,:\, z_l^* = k\}$ and $\xi m/K$ is the smallest cluster size. Assuming a mixture model with uniform mixing distributions, when $m$ is sufficiently large, $\xi \approx 1$. Note that the clusters are identifiable up to a change in label. Let $\Phi$ denote the set of all re-labelling function. The mis-clustering rate of an assignment $z$ is defined as
$$ \ell(z, z^*) = \min_{a \in \Phi} \frac{1}{m} \sum_{l=1}^m \ind[a(z_l) \neq z^*_l] \,.$$

\begin{theorem}[Theorem 3.2 of \cite{zhang2022leave}]\label{thm:spectral-clustering} Consider the generative model described above. Assume that $\epsilon_l \sim \SG(\sigma^2)$ independently with mean zero for $l\in [m]$. Asssume that $\beta m/K^4 \geq 400$. There exist constants $C, C', C_1, C_2$ such that under the assumption that
$$ \psi := \frac{\sqrt \xi \sqrt m \Delta}{K^2 \lVert E \rVert_{op} } > C\,, $$
and $\rho = \frac{T}{\lVert E\rVert_{op}}$ satisfies $C_1 \leq \rho \leq \frac{\psi}{C_2}$, then the output $z$ of Algorithm \ref{alg:spectral-clustering} satisfies
$$ \E \ell(z, z^*) \leq \exp{-(1- C'(\rho\psi^{-1}+\rho^{-1})) \frac{\Delta^2}{8\sigma^2} } + \exp{-\frac{m}{2}}\,.$$
If we further assume that $\psi, \rho \rightarrow \infty$ and $\rho/\psi = o(1)$, then
% and the threshold $T$ is appropriately chosen, then the output of the spectral clustering algorithm with adaptive dimension reduction, Algorithm (\ref{alg:spectral-clustering}), satisfies
$$ \E[\ell(\hat z, z^*)] \leq \exp{-(1-o(1))\frac{\Delta^2}{8\sigma^2}} + \exp{-m/2} \,.$$
\end{theorem}

\newpage
\section{Spectral Initialization}

This section is dedicated to proving Theorem \ref{thm:top-L-mixture-error-init}. At a high level, the proof consists of three main components. Firstly, we want to show that under the generative model described in Theorem \ref{thm:top-L-mixture-error-init}, the inter-center gap is large. This allows us to show that the mis-clustering error incurred by Algorithm \ref{alg:spectral-clustering} is small using Theorem \ref{thm:spectral-clustering}. The second component is the finite sample error bound for the least squares estimator in Algorithm \ref{alg:least-squares-estimate}. Lastly, we combine the two results to show that under the generative model in the theorem statement, the pairwise probability estimates obtained from the clusters produced by the spectral clustering algorithm are accurate and therefore Algorithm \ref{alg:least-squares-estimate} and, by extension, Algorithm \ref{alg:spectral-init} also return accurate parameter estimates.

\subsection{The Inter-center Gap}
Define $\kappa$ to be the dynamic range, i.e., $\kappa := \max_{i\in[n], k\in[K]} \, \lvert \theta_i^{*k}\rvert$. In our generative model, $\kappa \lesssim  \sqrt{\log n}$. The dynamic range is the reason why the term $\exp{O(\sqrt{\log n})}$ appears in our bound. The rest of this section is devoted to showing that the inter-center gap $\Delta$ is sufficiently large under said generative model. The following theorem shows that in the special case when $L = n$, the mixture model in Theorem \ref{thm:top-L-mixture-error-init} has a large inter-center gap (with high probability).
\begin{theorem}\label{thm:normal-gap} Fix a constant number of mixture components $K$. Suppose that for mixture component $k\in [K]$, ${\theta^*}^k \sim \N(0, I_n)$. For a sufficiently large $n$, there exist constants $C, c >0$ such that
$$ \Delta \geq \frac{c}{\exp{C\sqrt{\log n}}} \cdot \sqrt{n^2 - n\log n}  $$
with probability at least $1- O(\frac{K^2}{n^{10}}) - \exp{O(-n)}$.
\end{theorem}

\begin{proof} We restrict our analysis to just two mixture components $\theta$ and $\theta'$ for now. To obtain the high probability guarantee for $K$ mixture components, a simple union bound argument suffices.

The mapping function under the case Plackett-Luce is exactly the link function for the BTL model. Specifically, for items $i, j$ with partworths ${\theta^*_i}, {\theta^*_j}$ respectively,
$$ P_{ij} = \frac{1}{1 + \exp{-({\theta^*_i} - {\theta_j^*})}} \,.$$
The separation of the two pairwise centers satisfies
\begin{align*}
\Delta^2 &= \sum_{i < j} (P_{ij} - P'_{ij})^2\\
&\geq \sum_{i < j} l(\kappa)^2 \cdot \big(({\theta^*_i} - {\theta^*}) - ({\theta_i^*}' - {\theta_j^*}')\big)^2\,,
\end{align*}
where $l(\kappa)$ satisfies
$$\bigg\lvert\frac{1}{1 + \exp{-x}} - \frac{1}{1 + \exp{-y}} \bigg\rvert  \geq l(\kappa) \cdot \lvert x - y\rvert \,\forall \, x, y \in [-2\kappa, +2\kappa] \,.  $$
Clearly, $l(\kappa)$ is dependent on the dynamic range $\kappa$. Since the logit function is continuous everywhere, we can lower bound $l(\kappa)$ by lower bounding the gradient of the logit function within the dynamic range, which is
$$ \frac{\exp{-x}}{(1+\exp{-x})^2} $$
It is well known that for $\epsilon_1, \ldots, \epsilon_n \sim N(0, 1) $, there exists a constant $C$ such that $\max_i  \,\lvert \epsilon_i \rvert \leq 2C\sqrt{\log n}$ with probability at least $1 -\frac{1}{n^{10}}$. There also exists a constant $c' > 0$ such that
$$ l(\kappa) \geq \frac{\exp{-2C\sqrt{\log n}}}{(1+\exp{-2C\sqrt{\log n}})^2} \geq c'\,\exp{-2C\sqrt{\log n}} \,. $$
We will now tackle the remaining term in the expression for $\Delta^2$. We have
\begin{align*}
&\sum_{i < j} \big(({\theta_i^*} - {\theta_j^*}) - ({\theta_i^*}' - {\theta_j^*}')\big)^2\\
&= \frac{1}{2} \sum_{i \neq j}  \big(({\theta_i^*} - {\theta_j^*}) - ({\theta_i^*}' - {\theta_j^*}')\big)^2\\
&= \frac{1}{2} \sum_{i\neq j} \big( ({\theta_i^*} - {\theta_i^*}') - ({\theta_j^*} - {\theta_j^*}') \big)^2\\
&= \frac{1}{2} \sum_{i\neq j} ({\theta_i^*} - {\theta_i^*}')^2 + ({\theta_j^*} - {\theta_j^*}')^2 - 2({\theta_i^*} - {\theta_i^*}')({\theta_j^*} - {\theta_j^*}')\\
&= (n-1) \cdot \lVert {\theta^*} - {\theta^*}' \rVert_2^2 - ({\theta^*} - {\theta^*}')^\top (\mb 1_n \mb 1_n^\top - I_n) ({\theta^*} - {\theta^*}')\\
&= n \cdot \lVert {\theta^*} - {\theta^*}' \rVert_2^2 - ({\theta^*} - {\theta^*}')^\top (\mb 1_n \mb 1_n^\top) ({\theta^*} - {\theta^*}')  \\
&= n \cdot \lVert {\theta^*} - {\theta^*}' \rVert_2^2 - \big(({\theta^*} - {\theta^*}')^\top\mb 1_n\big)^2  \quad (*)\,. 
\end{align*}

~\\
For the second term, note that $\theta^* - {\theta^*}' \sim \calN(0, 2I_n)$. Invoking Theorem \ref{thm:concentrate-normal-sum}, there exists a constant $c''$ (e.g., 80) such that
$$ ({\theta^*} - {\theta^*}')^\top\mb 1_n \leq \sqrt{c''\,n\log n} $$
with probability ${n^{-10}}$.

~\\
As for the $\lVert \theta^* - {\theta^*}'\rVert_2$ term. Applying Lemma \ref{lem:concentrate-norm} gives
 $$ \lVert {\theta^*} - {\theta^*}' \rVert^2_2 \geq 0.9n $$
with probability $1 - O(\exp{-n})$.

~\\
Going back to the display $(*)$, we have
\begin{align*}
\Delta &\geq \exp{-C\sqrt{\log n}} \cdot \sqrt{0.9 n^2 - c''\, n\log n}
\end{align*}
with probability at least $1 - O(n^{-10}) - O(\exp{-n})$. Applying union bound over all pairs of mixture components in $[K]$ gives us the conclusion in the theorem statement.
\end{proof}

The following theorem generalizes the previous theorem to a general $L$.
\begin{theorem}\label{thm:gap-top-L} Fix a constant number of mixture components $K$. Suppose that for mixture component $k\in [K]$, $\theta_{1:L}^{*k} \sim \N(0, I_L)$ and $\theta_{L+1:n}^{*k}=0$.
There exist constants $C, c >0$ such that
$$ \Delta \geq \frac{c}{\exp{C\sqrt{\log n}}} \cdot \sqrt{nL - L\log n}  $$
with probability at least $1- O(\frac{K^2}{n^{10}}) - O(K^2\exp{-L})$.
\end{theorem}

\begin{proof} Following the same argument as in the proof of Theorem \ref{thm:normal-gap}, there exists a constant $c' > 0$ such that we decompose the gap between two cluster centers as
\begin{align*}
\Delta^2 &= \sum_{i < j} (P_{ij} - P'_{ij})^2\\
&\geq \sum_{i < j} l(\kappa)^2 \cdot \big(({\theta_i^*} - {\theta_j^*}) - ({\theta_i^*}' - {\theta_j^*}')\big)^2\,,\\
&= \bigg(\exp{-C\sqrt{\log n}}\bigg)^2 \cdot \bigg( n \, \lVert {\theta^*} - {\theta^*}' \rVert_2^2 - \big(({\theta^*} - {\theta^*}')^\top \mb 1_n\big)^2 \bigg)\,. 
\end{align*}
Note that by design,
$$ {\theta^*} - {\theta^*}' = [({\theta^*} -{\theta^*}')_{1:L}, 0, \ldots, 0] \,.$$
Applying Lemma \ref{lem:concentrate-norm}, we have
$$ \lVert {\theta^*} -{\theta^*}' \rVert_2  = \lVert ({\theta^*} -{\theta^*}')_{1:L} \rVert_2 \geq  \sqrt{0.9L} $$
with probability at least $1 - O(\exp{-L})$. 

~\\
For the second term, by Theorem \ref{thm:concentrate-normal-sum}, there exists a constant $c$ such that
$$  ({\theta^*} - {\theta^*}')^\top \mb 1_n = ({\theta^*} -{\theta^*}')_{1:L}^\top \mb 1_L < \sqrt{c\, L\log n}  $$
with probability at least $O(n^{-10})$. The rest of the proof proceeds similarly to that of Theorem \ref{thm:normal-gap}.
\end{proof}

\subsection{The Least Squares Estimator}
In this section, we obtain a general finite sample error guarantees for the least squares estimator (Algorithm \ref{alg:least-squares-estimate}). Recall that Algorithm \ref{alg:least-squares-estimate} apply a mapping function before solving a constrained least squares optimization problem. The mapping function used in the algorithm, the logit function, is specific to the Plackett-Luce/BTL model. If we change the mapping function to, say the inverse Normal CDF, we can apply Algorithm \ref{alg:least-squares-estimate} to parameter estimation under the Thurstone model \citep{thurstone2017law} (See Definition 1). We believe that these results can be extended to other ranking or paired comparison models, or to an incomplete observation setting where certain pairs of items are not compared.

\begin{lemma}\label{lem:least-squares-guarantee} Let $\hat\phi \in \R^{n\times n}$ be the transformed pairwise measurement in Algorithm \ref{alg:least-squares-estimate} and $\phi^*$ be its ideal value. Then the output $\theta$ of the algorithm satisfies
$$ \lVert N(\theta) - N(\theta^*) \rVert^2_2 \leq \frac{\lVert \hat\phi - \phi^*\rVert_F^2}{2n} \,.$$
\end{lemma}

\begin{proof} We first rewrite the least squares optimization problem in a more convenient form. Let $\vec(\phi^*)$ denote the vectorization of $\phi^*$ (row major order). There exists a matrix $Z \in \{0, 1, -1\}^{n(n-1)\times n}$ such that $\vec(\phi^*) = Z \theta^*$ for any $\theta^*$. Specifically, $Z = [Z_1, \ldots, Z_n]$ is a block matrix consisting of of $n$ blocks of dimension $n-1\times n$ as shown below.
$$\begin{bmatrix}
\phi^*_{12}\\ 
\phi^*_{13}\\ 
\ldots\\
\phi^*_{1n}\\
\phi^*_{21}\\ 
\ldots\\ 
\phi^*_{n,n-1}
\end{bmatrix} = \begin{bmatrix}
Z_1\\
Z_2\\
\ldots\\
Z_n \\
\end{bmatrix} \theta^* \,,$$
where
$$Z_1 = \begin{bmatrix}
1 &-1 &0 &0 &\ldots &0\\
1 &0 &-1 &0 &\ldots &0\\
1 &0 &0 &-1 &\ldots &0\\
\ldots &\ldots &\ldots &\ldots &\ldots &\ldots \\
1 &0 &0 &0 &\ldots &-1\\
\end{bmatrix} \in \R^{n-1\times n} \,,$$
$$Z_2 = \begin{bmatrix}
-1 &1 &0 &0 &\ldots &0\\
0 &1 &-1 &0 &\ldots &0\\
0 &1 &0 &-1 &\ldots &0\\
\ldots &\ldots &\ldots &\ldots &\ldots &\ldots \\
0 &1 &0 &0 &\ldots &-1\\
\end{bmatrix}  \in \R^{n-1\times n} \,,$$
and so on. One can see that the $Z_i$'s are different column-permuted versions of $Z_1$. Namely, there exist permutation matrices $R_1, \ldots, R_{n-1}$ such that $Z_2 = Z_1 R_1, Z_3 = Z_1 R_1 R_2$, .etc and $Z_n = Z_1 R_1\ldots R_{n-1}$.
We can also check that $Z_1 \mb 1_n = 0$.

Under our new notations, the least squares optimization problem in Algorithm \ref{alg:least-squares-estimate} is 
$$ \hat\theta = \arg\min_{\theta} \lVert \vec(\hat\phi) - Z\theta \rVert_2\,. $$
It is well known (cf., \cite{strang1993introduction}) that the least squares solution is
$$ \hat \theta = (Z^\top Z)^{\dagger}Z^\top \vec({\hat\phi}) \,, $$
where $(Z^\top Z)^{\dagger}$ is the pseudo-inverse of $Z^\top Z$. On the other hand, we also have
$$ \theta^* = (Z^\top Z)^{\dagger}Z^\top \vec({\phi^*}) \,.$$
Combining the above two equalities gives
$$ \hat\theta - \theta^* =  (Z^\top Z)^{\dagger}Z^\top (\vec(\hat\phi) - \vec(\phi^*)) \,. $$
% Hence,
% $$ \lVert \hat \theta - \theta^* \rVert_2^2 = (\vec(\hat\phi) - \vec(\phi^*))^\top \, Z((Z^\top Z)^{\dagger})^2Z^\top \, (\vec(\hat\phi) - \vec(\phi^*)) \,. $$
Now, since we care about the normalized $\ell_2$ norm between $\theta$ and $\theta^*$, we have
\begin{align*}
\lVert N(\hat \theta) - N(\theta^*)\rVert_2^2 &= \lVert \hat\theta - \frac{1}{n}\mb 1^\top \hat\theta \cdot \mb 1 - \theta^* + \frac{1}{n}\mb 1^\top \theta^*\cdot \mb 1\rVert_2^2\\
&= \lVert \hat\theta - \theta^* + (\frac{1}{n}\mb 1^\top \theta^* - \frac{1}{n}\mb 1^\top \hat\theta) \cdot \mb 1 \rVert_2^2\\
&= \lVert \hat\theta - \theta^* \rVert_2^2 - 2 \,\langle \frac{1}{n} \mb 1^\top (\hat\theta - \theta^*) \cdot \mb 1, \hat\theta - \theta^*\rangle + \langle \frac{1}{n}\mb 1^\top(\hat\theta - \theta^*)\cdot \mb 1,  \frac{1}{n}\mb 1^\top(\hat\theta - \theta^*) \cdot \mb 1\rangle\\
&= \lVert \hat\theta - \theta^* \rVert_2^2 - \frac{2}{n}\cdot ((\hat\theta - \theta^*)^\top \mb 1)^2 + \frac{1}{n^2}\cdot \mb 1^\top \mb 1 ((\hat\theta - \theta^*)^\top \mb 1)^2\\
&= \lVert \hat\theta - \theta^* \rVert_2^2 - \frac{1}{n}\cdot (\hat\theta - \theta^*)^\top \mb 1\mb 1^\top  (\hat\theta - \theta^*)\\
&= (\theta - \theta^*)^\top (I - \frac{\mb 1}{\sqrt n}\frac{\mb 1}{\sqrt n}) (\theta - \theta^*)\\
&= (\vec(\hat\phi) - \vec(\phi^*)) Z (Z^\top Z)^{\dagger} (I - \frac{\mb 1}{\sqrt n}\frac{\mb 1}{\sqrt n})(Z^\top Z)^{\dagger} Z^\top  (\vec(\hat\phi) - \vec(\phi^*))\,.
\end{align*}
Therefore,
\begin{align*}
\lVert N(\hat\theta) - N(\theta^*) \rVert_2^2 &\leq \lambda_{\max}\bigg( Z (Z^\top Z)^{\dagger} (I - \frac{\mb 1}{\sqrt n}\frac{\mb 1^\top}{\sqrt n})\underbrace{(Z^\top Z)^{\dagger} Z^\top}_{A^\top}  \bigg) \cdot \lVert \hat\phi -\phi^*\rVert_F^2 \\
&= \lambda_{\max}\bigg( A (I - \frac{\mb 1}{\sqrt n}\frac{\mb 1^\top}{\sqrt n})A^\top  \bigg) \cdot \lVert \hat\phi -\phi^*\rVert_F^2 \\
&= \lambda_{\max,\indep}\bigg(  A^\top A \bigg) \cdot \lVert \hat\phi -\phi^*\rVert_F^2 \\
&=  \lambda_{\max,\indep}\bigg((Z^\top Z)^{\dagger} Z^\top \, Z  (Z^\top Z)^{\dagger} \bigg) \cdot \lVert \hat\phi -\phi^*\rVert_F^2\\
&= \lambda_{\max,\indep}( (Z^\top Z)^{\dagger}) \cdot \lVert \hat\phi -\phi^*\rVert_F^2\,,
\end{align*}
where $\lambda_{\max, \indep}$ is the maximum eigenvalue associated with the eigenspace orthogonal to the vector $\mb 1_n$. 
% Before veryfing the second step of the above derivation, let us continue with the rest of the proof. 
% The rest of the proof is devoted to obtaining an upper bound on $\lambda_{\max, \indep}\bigg( Z((Z^\top Z)^{\dagger})^2Z^\top\bigg)$.
% where $\lambda_{\max}$ denotes the maximum eigenvalue. Let $Z = USV^\top$ be the singular value decomposition of $Z$. We have
% \begin{align*}
% Z((Z^\top Z)^{\dagger})^2Z^\top &= USV^\top (VS^2V^\top)^{\dagger} \,  (VS^2V^\top)^{\dagger} VSU^\top\\
% &= USV^\top V {S^{-2}} V^\top V S^{-2} V^\top VSU^\top \,\\
% &= US {S^{-4}} SU^\top \,,
% \end{align*}
% where, by definition of the pseudo-inverse, we define the inverse of the diagonal matrix $S$ as 
% $$({S^{-4}})_{ii} = \begin{cases} 0 &\text{if } S_{ii} = 0 \\ S_{ii}^{-4} &\text{otherwise} \end{cases}\,. $$
% Then, with $S^{-2}$ defined similarly to $S^{-4}$ above, we have
% $$ Z((Z^\top Z)^{\dagger})^2Z^\top = U {S^{-2}} U^\top\,. $$
% Continuing with our bound earlier, we have
% \begin{align*}
% &\lambda_{\max}\bigg( Z (Z^\top Z)^{\dagger} (I - \frac{\mb 1}{\sqrt n}\frac{\mb 1}{\sqrt n})(Z^\top Z)^{\dagger} Z^\top  \bigg)\\
% &= \lambda_{\max}\bigg( US^{-1} (I - \frac{\mb 1}{\sqrt n}\frac{\mb 1}{\sqrt n}) S^{-1}U^\top \bigg)\\
% &= \lambda^{-1}_{\min, \indep} \bigg( Z^\top Z) \,,
% \end{align*}
% where $\lambda_{\min,\indep}$ is the minimum eigenvalue associated with the eigenspace orthogonal to $\mb 1$.
% ~\\
% We need to analyze the diagonal entries of $S^{-1}$. For that, it suffices to analyze $S$. One can verify the following.
One can verify the following.
\begin{align*}
Z^\top Z &= Z_1^\top Z_1 + Z_2^\top Z_2 + \ldots + Z_n^\top Z_n\\
&= \begin{bmatrix}
n-1 & -1 & -1 & \ldots & -1 \\ 
-1 & 1 & 0 & \ldots & 0 \\ 
-1 & 0 & 1 & \ldots & 0 \\ 
\vdots &  &  &  & \\ 
-1 & 0 & 0 & \ldots &1 
\end{bmatrix} + 
\begin{bmatrix}
1 & -1 & 0 & \ldots & 0 \\ 
-1 & n-1 & -1 & \ldots & -1 \\ 
0 & -1 & 1 & \ldots & 0 \\ 
\vdots &  &  &  & \\ 
0 & -1 & 0 & \ldots &1 
\end{bmatrix} +
\ldots
\begin{bmatrix}
1 & 0 & 0 & \ldots & -1 \\ 
0 & 1 & 0 & \ldots & -1 \\ 
0 & 0 & 1 & \ldots & -1 \\ 
\vdots &  &  &  & \\ 
-1 & -1 & -1 & \ldots &n-1 
\end{bmatrix} \\
&= 2n \, I_n - 2n\, \frac{\mb 1}{\sqrt n} \frac{\mb 1^\top}{\sqrt n}\,.
\end{align*}
Hence, $\lambda_{\min,\indep}(Z^\top Z) = 2n$ and $\lambda_{\max,\indep}((Z^\top Z)^{\dagger}) = \frac{1}{2n}$. This directly yields the inequality in the theorem statement.
\end{proof}

\subsection{Proof of Theorem \ref{thm:top-L-mixture-error-init}}
Let us introduce some notations. Let $C^{*k} = \{ l \in [m] \,\lvert \, z^*_l = k \}$ for $k\in [K]$ denote the true cluster assignment. To avoid cluttering the notation, let us suppress the relabeling operator here and consider $\hat C^k = \{l \in [m] \,\lvert \, z_l = k \}$ to be the cluster assignment produced by Algorithm \ref{alg:spectral-clustering} corresponding to mixture component $k$.

As noted before, when the mixing probabilities are uniform, $\beta^* = \frac{1}{K}\mb 1$, $\xi \approx 1$ and the size of each true cluster is $\approx \frac{m}{K}$ whp. In fact, for a cluster $k$, Hoeffding's inequality gives
$$ \Pr\bigg(\bigg\lvert \lvert C^{*k}\rvert - \frac{m}{K} \bigg\rvert > t \bigg) \leq 2 \,\exp{-\frac{2t^2}{m}} \,. $$
Hence, $\frac{0.9m}{K} \leq \lvert C^{*k} \rvert \leq \frac{1.1m}{K} $ with probability at least $1 - O(\exp{-m/K^2})$. Applying union bound over all clusters, we conclude that the event
$$ \calE_0 = \bigg\{ \lvert C^{*k} \rvert \in \bigg[\frac{0.9m}{K}, \frac{1.1m}{K}\bigg] \, \forall k \in [K] \bigg\} $$
happens with probability at least $1 - O(K\exp{-m/K^2})$.

~\\
\textbf{On the choice of $T$.} Theorem \ref{lem:operator-norm-bound} gives $\lVert E \rVert_{op} \lesssim \sqrt{n}({\sqrt m + \sqrt n})$ with probability at least $1-\frac{1}{(n+m)^{10}}$. The technical condition in \citet{zhang2022leave} sugggests that $T$ is at least a constant factor larger than $\lVert E \rVert$. It therefore suffices to pick $T = \sqrt{n}\sqrt{m+n} \sqrt{\log n}$. Additionally, the second bound in Theorem \ref{lem:operator-norm-bound} holds if the following event holds
$$ \calE'_0 = \bigg\{ \frac{\sqrt{\log n} \, K^2 \lVert E\rVert_{op} }{\xi\sqrt m \Delta} = o(1) \bigg\}\,. $$
We will show that this is satisfied under the generative model described in Theorem \ref{thm:top-L-mixture-error-init}. For now, let us define the following events.
$$ \calE_1 = \bigg\{ \Delta \geq \frac{\sqrt d}{\exp{D/2\sqrt{\log n}}} \sqrt L \sqrt{n-\log n} \bigg\} \cup \calE_0 \cup \calE_0'\,. $$
$$ \calE_2 = \bigg\{ \Delta \geq \frac{\sqrt d}{\exp{D/2\sqrt{\log n}}} \sqrt {n^{\alpha}} \sqrt{n-\log n} \bigg\} \cup \calE_0  \cup \calE_0'\,. $$
$d$ and $D$ are the constants in Theorem \ref{thm:gap-top-L}. Recall the sub-gaussian norm of the pairwise representation of permutations in Lemma \ref{lem:sub-gaussian-norm-general}. For the rest of the proof, we use the same $\tau$.

\begin{lemma}\label{lem:clustering-guarantee} Suppose that event $\calE_1$ or event $\calE_2$ holds and $m\geq c K^4$ or a sufficiently large constant $c$. Then the clustering $\hat z$ output of Algorithm \ref{alg:spectral-clustering} satisfies the following.
\begin{align*}
&\Pr\bigg( m \cdot \ell(\hat z, z^*)  > \frac{m}{K}\,\exp{-\frac{\Delta^2}{16\tau n}} \,\bigg\lvert \, \calE_1 \bigg)\\
&\leq K\cdot \exp{-(1- o(1)) \frac{\Delta^2}{16\tau n} } + K \cdot \exp{-\frac{m}{2} + \frac{\Delta^2}{16\tau n}}  \,,
\end{align*}
\end{lemma}

\begin{proof} Under the assumptions stated in the lemma, the output of Algorithm \ref{alg:spectral-clustering} achieves exponentially decaying mis-clustering error rate and
\begin{align*}
&\Pr\bigg( m\cdot \ell(z, z^*) > t \,\bigg\lvert \, \calE_1 \bigg) =  \Pr\bigg(\ell(z, z^*) > \frac{t}{m} \bigg) \\
&\leq\frac{m}{t} \cdot \exp{-(1- o(1)) \frac{\Delta^2}{8\sigma^2} } + \frac{m}{t} \cdot \exp{-\frac{m}{2}} \,,
\end{align*}
where the last step comes from applying Markov's inequality and $\psi, \rho, C'$ are defined in Theorem \ref{thm:spectral-clustering} and $\sigma^2 = \tau n$. The theorem statement is obtained by replacing $t$ with $\frac{m}{K} \cdot \exp{-\frac{\Delta^2}{16\tau n}}$.
\end{proof}

\begin{reptheorem}{thm:top-L-mixture-error-init} Consider a mixture of $K$ Plackett-Luce models with uniform mixing probabilities. Suppose that $\theta^{*k}_{i} = 0 \,\forall i \in [L+1:n]$ and $\theta^{*k}_{1:L} \sim \N(0, I_L)$ for $k\in [K]$. Fix a constant $\alpha > 0$. There exist constants $c, c_1, C_1, C_2$ such that if $m\geq c\max \{K^4, Kn\}$, the output $\hat\btheta$ of Algorithm \ref{alg:spectral-init} satisfies the following. 
If $L \geq c_1\,\exp{C_1 \sqrt{\log n}}$,
\begin{equation*}
% \begin{multline*} 
\text{dist}(\hat\btheta, \btheta^*) = O\bigg(\exp{O(\sqrt{\log n})}\cdot \bigg(\sqrt{\frac{K^2n\log n}{m}} + \frac{\sqrt{Kn}}{e^{ L^{0.99}}} \bigg)  \bigg)\,
% \end{multline*}
\end{equation*}
with probability $1- O(\frac{K}{n^8}) + O(K^2n^2\exp{-L^{0.99}})$. 
If $L \geq C_2 n^\alpha$ and assuming that $n = \omega(\log m)$,
\begin{equation*}
    \text{dist}(\hat\btheta, \btheta^*) = O\bigg( \exp{O(\sqrt{\log n})}\, \sqrt{\frac{K^2n\log n}{m}}\, \bigg)\,
\end{equation*}
with probability $1- O(\frac{K}{n^8}) + O(K^2n^2\exp{-n^{\alpha}})$.
\end{reptheorem}

\begin{proof} Since the output of Algorithm \ref{alg:spectral-clustering} are $K$ clusters which are then used to estimate the pairwise preference probabilities -- the input to Algorithm \ref{alg:least-squares-estimate} -- it suffices to show that each pairwise center $\hat P^k$ is close to the corresponding true centers $P^{*k}$. 

Focusing on a cluster $k \in [K]$ and a single pair $(i, j)$, let $N^k_{ij}$ denote the number of times that item $j$ `beats' item $i$ among the permutations in cluster $\hat C^k$.
\begin{align*}
N^k_{ij} &= \sum_{l\in  C^{*k}} \ind [\pi_l(i) < \pi_l(j)] - \sum_{l\in C^{*k}, l\notin \hat C^k} \ind [\pi_l(i) < \pi_l(j)] + \sum_{l \notin C^{*k}, l\in \hat C^k} \ind [\pi_l(i) < \pi_l(j)]\,.
% &\asymp m P_{ji}^* \pm O(\sqrt{m\log n}) \pm m \cdot O\bigg( \exp{-\frac{\Delta^2}{\sigma^2}} + \exp{-m} \bigg)
\end{align*}
We will analyze these terms separately. By Hoeffding's inequality, the first term can be bounded as
\begin{align*}
\Pr\bigg(\bigg\lvert \sum_{l\in C^{*k}} \ind [\pi_l(i) < \pi_l(j)] - \lvert C^{*k} \rvert \cdot P_{ij}^*\bigg\lvert > t \bigg) \leq 2\exp{-\frac{2t^2}{ \lvert C^{*k}\rvert }} \leq  2\exp{-\frac{2K\,t^2}{ 1.1m}}  \,.
\end{align*}
Therefore, there exists a constant $d_1$ such that
$$\big\lvert \sum_{l\in  C^{*k}} \ind [\pi_l(i) < \pi_l(j)] - \lvert  C^{*k} \rvert \cdot P_{ij}^*\big\lvert  \leq d_1 \sqrt{m/K \, \log n} $$
with probability at least $1-n^{-10}$. 

% By union bound, above bound holds \emph{for all pairs} with probability at least $1-n^{-8}$. Applying union bound over all clusters $k \in [K]$, we have the above bound hold for all pairs and clusters with probability at least $1-Kn^{-8}$.

~\\
For the other two terms in $N^k_{ij}$, note that by definition,
\begin{align*}
\bigg\lvert \sum_{l\in C^{*k}, l\notin \hat C^k} \ind [\pi_l(i) < \pi_l(j)] \bigg\rvert + \bigg\lvert \sum_{l \notin C^{*k}, l\in \hat C^k} \ind [\pi_l(i) < \pi_l(j)] \bigg\rvert \leq m \cdot \ell(z, z^*)\,.
\end{align*}
At this point, we consider the two regimes of $L$ separately as the gap $\Delta$ differs substantially between the two regimes, leading to different bounds on the quantity above.

~\\
\textbf{For the regime $L \geq c_1\exp{C_1\sqrt{\log n}}$}. By Theorem \ref{thm:gap-top-L}, the gap $\Delta$ satisfies the following, for some constants $d, D$.
$$ \Delta^2 \geq \frac{d}{\exp{D\sqrt{\log n}}} L(n - \log n) $$
with probability at least $1- \frac{K^2}{n^{10}} - O(K^2\exp{-L})$. Because $\exp{\sqrt{\log n}} = \omega(\log n)$, the ratio $\rho/\psi$ in Theorem \ref{thm:spectral-clustering} satisfies
$$ \frac{\sqrt{\log n} \, K^2 \lVert E\rVert_{op} }{\xi\sqrt m \Delta} \asymp \frac{\sqrt{\log n} \, K^2 \sqrt n \sqrt{m+n} }{ \,\exp{(C_1-D)\sqrt{\log n}}\, \sqrt{n - \log n}\sqrt m } = o(1)\,.$$ 

Then by Theorem \ref{thm:gap-top-L}, Lemma \ref{lem:operator-norm-bound} and considering the generative model in the theorem statement, event $\calE_1$ happens with probability at least 
$1- \frac{K^2}{n^{10}} - O(K^2\exp{-L}) - O(K\exp{-m/K^2}) - \frac{1}{(n+m)^{10}}$. The last two terms in the probability guarantee diminish much faster than the other terms so we can simplify as
$$ \Pr(\calE_1) \geq 1 - O(\frac{K^2}{n^{10}}) - O(K^2\exp{-L}) \,.$$
By Lemma \ref{lem:clustering-guarantee},
\begin{align*} 
&\Pr\bigg( m \cdot \ell(\hat z, z^*) > \frac{m}{K}\,\exp{-\frac{\Delta^2}{16\tau n}} \,\bigg\lvert \, \calE_1 \bigg) \\
&\leq K\cdot \exp{-(1- o(1)) \frac{\Delta^2}{16\tau n} } + K \cdot \exp{-\frac{m}{2} + \frac{\Delta^2}{16\tau n}}\\
&= K\cdot \exp{-\frac{(1- o(1)) d}{16\tau} \cdot \frac{L(n-\log n) }{\exp{D\sqrt{\log n}} \, n} } + K \cdot \exp{-\frac{m}{2} + \frac{d}{16\tau} \cdot \frac{L(n-\log n) }{\exp{D\sqrt{\log n}} \, n}}\\
&= K\cdot \exp{- \frac{(1- o(1)) d}{16\tau} \cdot \frac{L}{\exp{D\sqrt{\log n}}} } + K \cdot \exp{-\frac{m}{2} + \frac{d(1-o(1))}{16\tau} \cdot \frac{L}{\exp{D\sqrt{\log n}}}}
\end{align*}
There exists a constant $c$ such that if $m \geq c n$, then the second exponential term is bounded as
$$  K \cdot \exp{-\frac{m}{2} + \frac{d(1-o(1))}{16\tau} \cdot \frac{L}{\exp{D\sqrt{\log n}}}} \leq K \exp{-n}\,, $$
which is dominated by the first exponential term. Hence, we can simplify the probability bound on the mis-clustering error as
\begin{align*} 
&\Pr\bigg( m \cdot \ell(\hat z, z^*) > \frac{m}{K}\,\exp{-\frac{\Delta^2}{16\tau n}} \,\bigg\lvert \, \calE_1 \bigg)\\
& \leq 2K \cdot\exp{- \frac{(1- o(1))d}{16\tau} \cdot \frac{L}{\exp{D\sqrt{\log n}}} }  \\
&= 2K \cdot\exp{- \frac{(1- o(1))d c_1}{16\tau} \cdot \exp{(C_1 - D)\sqrt{\log n}}} \\
&[\text{There exist values for constants $c'_1, C'_1$ such that if} \\
&\text{$c_1 > c_1', C_1 > C_1'$ then the above term is bounded by}]\\
&\leq 2K \cdot\exp{- c_1^{0.99}\exp{0.99C_1 \sqrt{\log n}}} \\
&\leq 2K \cdot \exp{-L^{0.99}}
\end{align*}
% There also exist constants $c''_1, C''_1$, such that if $L \geq c_1'' \exp{C''_1\sqrt{\log n}}$ then $\frac{\Delta^2}{16\tau n} \asymp \exp{(C_1-D)\sqrt{\log n}} \gtrsim \exp{L^{0.99}} \geq 10\log n $. Set $c_1 = \max\{c_1', c_1''\}$ and $C_1= \max\{C_1', C_1''\}$.

~\\
Conditioned on event $\calE_1$ and with the choice of $c_1, C_1$ above, the following happens with probability at least $1- 2K \cdot \exp{-L^{0.99}} - \frac{1}{n^{10}}$ for a single pair $i,j$ and cluster $k$.
$$ \bigg\lvert N_{ij} - \frac{m}{K} \cdot P_{ij}^{*k} \bigg\rvert \lesssim \sqrt{m/K \, \log n} + m \ell(z, z^{*}) \asymp \sqrt{m/K \, \log n}  + \frac{m}{Ke^{L^{0.99}}} \,.$$
Since $ C^{*k} \asymp \frac{m}{K}\, $ and $ \lvert \hat C^k - C^{*k} \rvert \leq  \frac{m}{Ke^{L^{0.99}}}\,, $ we have $\hat C^k \asymp \frac{m}{K}$.
Consequently,
\begin{align*}
\lvert \hat P^k_{ij} - P^{*k}_{ij} \rvert =\bigg\lvert \frac{N^k_{ij}}{\lvert \hat C^k\rvert} - P_{ij}^* \bigg\rvert \asymp \frac{1}{m/K} \cdot \bigg\lvert N^k_{ij} - {m/K} \cdot {P^*_{ij}}  \bigg\rvert  \lesssim \frac{\sqrt K\,\sqrt{\log n}}{\sqrt m} + \frac{1}{e^{L^{0.99}}} \,.
\end{align*}
We apply union bound over all pairs $i\neq j$. Conditioned on event $\calE_1$, with probability at least $1- \frac{2Kn^2}{\exp{L^{0.99}}}-\frac{1}{n^8}$.
$$ \lvert \hat P^k_{ij} - P^{*k}_{ij} \rvert \lesssim  \frac{\sqrt K \, \sqrt{\log n}}{\sqrt m} + \frac{1}{e^{L^{0.99}}} $$
for all pairs $i, j$ in cluster $k$. Apply a union bound argument over all clusters, the above bound holds for all $k\in [K]$ and all pairs with probability at least $1- \frac{2K^2n^2}{\exp{L^{0.99}}} - \frac{K}{n^8}$.
Since the event $\calE_1$ happens with probability at least $1- O(\frac{K^2}{n^{10}}) - O(K^2\exp{-L})$, we can show that
\begin{align*} 
&\Pr\bigg(\lvert \hat P^k_{ij} - P^{*k}_{ij} \rvert \gtrsim  \frac{\sqrt K \, \sqrt{\log n}}{\sqrt m} + \frac{1}{e^{L^{0.99}}} \quad\forall k\in [K], i\neq j\in [n] \bigg)\\
&\leq \Pr(\sim \calE_1) + \Pr(\calE_1) \cdot \Pr\bigg(\lvert \hat P^k_{ij} - P^{*k}_{ij} \rvert \gtrsim  \frac{\sqrt K \, \sqrt{\log }}{\sqrt m} + \frac{1}{e^{L^{0.99}}} \quad\forall k\in [K], i\neq j\in [n] \,\bigg\lvert\, \calE_1 \bigg)\\
&\leq O(\frac{K^2}{n^{10}}) + O(K^2\exp{-L}) \\
& + (1 - O(\frac{K^2}{n^{10}}) - O(K^2\exp{-L})) \cdot \bigg(\frac{2K^2n^2}{\exp{L^{0.99}}} + \frac{K}{n^8}\bigg)\\
&\leq O(\frac{K}{n^8}) + O(K^2n^2\exp{-L^{0.99}})\,.
\end{align*}

~\\
\textbf{For the regime $L \geq C_2 n^{\alpha}$.} Again by Theorem \ref{thm:normal-gap},
$$ \Delta^2 \geq \frac{d}{\exp{D\sqrt{\log n}}} \cdot (nL - L\log n) \asymp \frac{dC_2n^\alpha}{\exp{D\sqrt{\log n}}} \cdot C_2n^\alpha \cdot (n - \log n) \, $$
with probability at least $1 - \frac{K^2}{n^{10}} - O(K^2\exp{-C_2n^{\alpha}})$.
One can check that $\rho/\psi = o(1)$ since $\exp{D\sqrt{\log n}} = o(n^{\alpha}) $ for any $\alpha > 0$.
Therefore, event $\calE_2$ happens with probability at least $1- \frac{K^2}{n^{10}} - O(K^2\exp{-C_2n^{\alpha}}) - O(K\exp{-m/K^2}) - \frac{1}{(n+m)^{10}} $.
Following a similar argument as in the previous regime of $L$, we have
\begin{align*} 
&\Pr\bigg( m \cdot \ell(\hat z, z^*) > \frac{m}{K}\,\exp{-\frac{\Delta^2}{16\tau n}} \,\bigg\lvert \, \calE_2 \bigg) \\
&\leq 2K\cdot \exp{-(1- o(1)) \frac{\Delta^2}{16\tau n} }\\ 
&= 2K\cdot \exp{- \frac{(1- o(1))d}{16\tau} \cdot \frac{L(n-\log n) }{\exp{D\sqrt{\log n}} \, n} }\\
&= 2K\cdot \exp{-\frac{(1- o(1))d}{16\tau} \cdot \frac{L}{\exp{D\sqrt{\log n}}}} \\
&[\text{There exist constants $C_2'$ such that if $C_2 > C_2'$ then}\\
&\text{the above term can be bounded as}]\\
&\leq 2K \exp{-n^{\alpha}}
\end{align*}
By the assumption that $n = \omega(\log m)$, there also exist constants $C''_2$, such that if $C_2 > C_2''$ then $\frac{\Delta^2}{16\tau n} \asymp \exp{n^{\alpha}} \geq \sqrt{m} $ and $\exp{-\frac{\Delta^2}{16\tau n}} \leq \frac{1}{\sqrt{m}} $. Set $C_2= \max\{C_2', C_2''\}$.
~\\
Conditioned on event $\calE_2$ and with the choice of $C_2$ above, the following happens with probability at least $1 - 2K\exp{-n^{\alpha}} - \frac{1}{n^{10}}$ for a single pair $i,j$ and cluster $k$.
$$ \bigg\lvert N_{ij} - \frac{m}{K} \cdot P_{ij}^{*k} \bigg\rvert \lesssim \sqrt{m/K \, \log n} + m \ell(z, z^{*}) \asymp \sqrt{m/K \, \log n}  + \sqrt{m/K} \,.$$
Clearly the first term dominates. Since $ C^{*k} \asymp \frac{m}{K}\, $ and $ \lvert \hat C^k - C^{*k} \rvert \leq  \sqrt{\frac{m}{K}} \,, $ we have $\hat C^k \asymp \frac{m}{K}$. Consequently,
\begin{align*}
\lvert \hat P^k_{ij} - P^{*k}_{ij} \rvert =\bigg\lvert \frac{N^k_{ij}}{\lvert \hat C^k\rvert} - P_{ij}^* \bigg\rvert \asymp \frac{1}{m/K} \cdot \bigg\lvert N^k_{ij} - {m/K} \cdot {P^*_{ij}}  \bigg\rvert  \lesssim \frac{\sqrt K\,\sqrt{\log n}}{\sqrt m} \,.
\end{align*}
We apply union bound over all pairs $i\neq j$ and cluster $k\in [K]$. Conditioned on event $\calE_2$, with probability at least $1- \frac{2K^2n^2}{\exp{-n^{\alpha}}}-\frac{K}{n^8}$, the following holds for all pairs in all cluster components.
$$ \lvert \hat P^k_{ij} - P^{*k}_{ij} \rvert \lesssim  \frac{\sqrt K \, \sqrt{\log n}}{\sqrt m} \,.$$
Since event $\calE_2$ happens with probability at least $1- O(\frac{K^2}{n^{10}}) - O(K^2\exp{-C_2n^{\alpha}})$, we have
\begin{align*} 
&\Pr\bigg(\lvert \hat P^k_{ij} - P^{*k}_{ij} \rvert \gtrsim  \frac{\sqrt K \, \sqrt{\log n}}{\sqrt m} \quad\forall k\in [K], i\neq j\in [n] \bigg)\\
&\leq \Pr(\sim \calE_2) + \Pr(\calE_2) \cdot \Pr\bigg(\lvert \hat P^k_{ij} - P^{*k}_{ij} \rvert \gtrsim  \frac{\sqrt K \, \sqrt{\log n}}{\sqrt m}  \quad\forall k\in [K], i\neq j\in [n] \,\bigg\lvert\, \calE_2 \bigg)\\
&\leq O(\frac{K^2}{n^{10}}) + O(K^2\exp{-n^{\alpha}}) \\
& + (1 -O(\frac{K^2}{n^{10}}) - O(K^2\exp{-C_2n^{\alpha}})) \cdot\bigg(\frac{2K^2n^2}{\exp{-n^{\alpha}}}+\frac{K}{n^8} \bigg)\\
&= O(\frac{K}{n^8}) + O(K^2n^2\exp{-n^{\alpha}})\,.
\end{align*}

~\\
\textbf{Bounding the Error introduced by the Logit Transformation.} From Lemma \ref{lem:least-squares-guarantee}, the output $\hat\theta$ satisfies
$$ \lVert \hat\theta^k - N(\theta^{*k}) \rVert_2 \leq \frac{\lVert \hat\phi^k - \phi^{*k} \rVert_F}{\sqrt{2n}} \,. $$
We have already obtain bound on $\lvert \hat P^k_{ij} - P^{*k}_{ij}\rvert $. We still need to account for the error introduced by the logit mapping function. By the sub-gaussian property, there exists a constant $B'$ such that $\lvert \theta_i \rvert \leq B'\sqrt{\log n} \,\,\forall i\in[n]$ with probability at least $1- \frac{1}{n^{8}}$. Then for $B = 2B'$, we have for all pairs $(i, j)$,
$$ P^*_{ij} \geq \exp{-2B\sqrt{\log n}} \,.$$
Recall that $\exp{\sqrt{\log n}} = \frac{\sqrt{\log n}}{\sqrt n}$ and $\exp{\sqrt{\log n}} = o\big( \exp{\exp{\sqrt{\log n}}} \big) $. Therefore, there exist constants $c, c'''_1, C'''_1$ such that if $m \geq cKn$ and $L \geq c'''_1 \exp{C'''_1\sqrt{\log n}}$, the following holds for all pairs $(i,j)$ and clusters $k \in [K]$.
$$\lvert \hat P^k_{ij} - P^{*k}_{ij} \rvert \lesssim \frac{\sqrt{K\log n}}{\sqrt m} + \frac{1}{e^{\sqrt L}} \leq \frac{1}{2} \exp{-2B\sqrt{\log n}}\, $$
with probability at least $1-O(\frac{K}{n^8})-O(\frac{K^2n^2}{\exp{L^{0.99}}})$. In other words,
$$ \hat P^k_{ij} \geq \frac{1}{2} \exp{-2B\sqrt{\log n}} \quad \forall i\neq j \in [n], k\in [K]\,. $$
Recall the logit mapping function $\phi(\hat p) = \ln\bigg(\frac{\hat p}{1-\hat p}\bigg)$. Its gradient is
$$  \frac{1}{\hat p} + \frac{1}{1-\hat p}\,. $$
Since the logit function is continuous, an upper bound on its function also suffices as a Lipschitz constant. Namely, for any $p, p'\in [\frac{1}{2} \exp{-2B\sqrt{\log n}}, 1-\frac{1}{2} \exp{-2B\sqrt{\log n}}]$,
$$ \bigg\lvert \ln\bigg(p/(1-p)\bigg) - \ln\bigg(p'/(1-p') \bigg) \bigg\rvert \leq 4\,\exp{2B\sqrt{\log n}}  \lvert p - p'\rvert \,. $$

~\\
\textbf{Putting things together.} For the small $L$ regime, we have with probability $1- O(\frac{K}{n^8}) - O(K^2n^2\exp{-L^{0.99}}) $, the following holds for all $k\in [K]$.
\begin{align*}
\lVert \hat\theta^k - \theta^{*k}\rVert_2^2 &\leq \frac{\lVert \hat\phi^k - \phi^{*k} \rVert_F^2}{2n}\\
&= \frac{\sum_{i, j} (\hat\phi^k_{ij} - \phi^{*k}_{ij} )^2 }{2n}\\\
&\lesssim \frac{\sum_{ij} \exp{4B\sqrt{\log n}} \cdot (\hat P^k_{ij} - P^{*k}_{ij} )^2 }{n}\\
&\lesssim  \exp{4B\sqrt{\log n}} n \cdot \bigg( \frac{\sqrt K \, \sqrt{\log n}}{\sqrt m} + \frac{1}{e^{L^{0.99}}} \bigg)
\end{align*}
For the large $L$ regime, we have with probability $1- O(\frac{K}{n^8}) - O(K^2n^2\exp{-n^{\alpha}}) $, the following holds for all $k\in [K]$.
\begin{align*}
\lVert \hat\theta^k - \theta^{*k}\rVert_2^2 &\leq \frac{\lVert \hat\phi^k - \phi^{*k} \rVert_F^2}{2n}\\
&= \frac{\sum_{i, j} (\hat\phi^k_{ij} - \phi^{*k}_{ij} )^2 }{2n}\\\
&\lesssim \frac{\sum_{ij} \exp{4B\sqrt{\log n}} \cdot (\hat P^k_{ij} - P^{*k}_{ij} )^2 }{n}\\
&\lesssim  \exp{4B\sqrt{\log n}} n \cdot \bigg( \frac{\sqrt K \, \sqrt{\log n}}{\sqrt m} \bigg)\\
\end{align*}
By noting that
$$ \lVert \btheta - \btheta^* \rVert_F \leq \sqrt K \cdot \max_{k} \, \lVert \theta^k - \theta^{*k}\rVert_2\,, $$
we obtain the bounds in the theorem statement. This finishes the proof.
\end{proof}

~\\
\textbf{Technical Remarks.} As noted earlier, the $\exp{O(\sqrt{\log n})}$ in the $\Delta$ quantity is related to the dynamic gap which scales as $\sqrt{\log n}$ in our generative model. However, the reader might see that this term appears in the final bound as a consequence of a worst case analysis. For this reason, we are not able to obtain finite sample error bound for the very sparse regime $L \asymp \log n$ or $L = O(1)$. However, it might be possible to obtain a more refined lower bound for $\Delta$ that removes this term. With this, one could extend the theorem to the small $L$ regime.

~\\
\textbf{LSR versus Least Squares for Initialization.} The reader might wonder why the spectral initialization does not use the LSR algorithm after the initial clustering step. This is a reasonable question since the algorithm is essentially a cluster-then-learn algorithm so the learning component can be delegated to the LSR algorithm. On the theoretical side, the least squares estimator presents an elegant theoretical analysis that can also be extended to more general RUMs other than PL. We propsoe the spectral initialization as a flexible algorithm for learning mixtures of general RUMs, not just mixtures of PLs. On the practical side, the least squares algorithm reduces the estimation problem to a constrained least-squares optimization problem which can be efficiently solved using off-the-shelf optimizers. Therefore, there are important reasons to consider applying the least squares estimator to the spectral initialization algorithm over the (unweighted) LSR algorithm. Experimentally, we observe that spectral initialization with LSR produces similar estimates to spectral initialization with the least squares estimator and the final estimates obtained by the EM algorithm are very similar.

\newpage
\section{The Weighted LSR Algorithm}
In this section, we will prove that the output of the weighted LSR algorithm is also the weighted MLE -- Equation (\ref{eqn:weighted-log-likelihood}).

% \duc{Note we change the Markov chain symbol to $M$}

~\\
\textbf{The Weighted Log-likelihood.} The connection between the weighted ILSR algorithm and the weighted log-likelihood estimate stems from the generative model of the PL model as a sequence of choice. Namely, for a ranking $\pi$. The log-likelihood decomposes as
$$ \log \Pr^{PL}(\pi\,\lvert\,\theta) = \sum_{i=1}^{n-1} \log\bigg( \frac{e^{\theta_{\pi_i}}}{\sum_{j\geq i} e^{\theta_{\pi_j}} }  \bigg)\,. $$

Given a ranking dataset $\Pi$ and sample weight vector $q$, the weighted log-likelihood is defined as
\begin{equation*}
\mathcal{L}(\Pi, q, \theta) = \sum_{l=1}^m \sum_{i=1}^{n-1} q_l \cdot \log\bigg( \frac{e^{\theta_{\pi_i}}}{\sum_{j\geq i} e^{\theta_{\pi_j}} }  \bigg)\,.
\end{equation*}
Similarly to the analysis of the unweighted LSR algorithm in \citet{maystre2015fast}, we draw the connection between MLE and spectral ranking via the following reparametrization. Let $\hat \theta$ be the weighted MLE. Now define
$$ \hat p_i = \frac{e^{\hat\theta_i}}{\sum_{k\in[n]} e^{\hat\theta_k}  }  \,\, i \in [n]\,.$$
Naturally, $\hat p \in \calP_n$ where $\calP_n$ is the open probability symplex in $n$ dimension.
In terms of $\hat p$, the weighted log-likelihood is
\begin{align*}
\mathcal{L}(\Pi, q, \hat p) = \sum_{l=1}^m \sum_{i=1}^{n-1} q_l \cdot \log\bigg( \frac{\hat p_{\pi_i}}{\sum_{j\geq i} \hat p_{\pi_j} }  \bigg)\,.
\end{align*}

~\\
\textbf{Choice Breaking.} Recall that the weighted ILSR algorithm first performs choice breaking on the permutations and the notion of choice breaking as a collection of choice enumerations composing the rankings. The weighted log-likelihood can be written in terms of the choice breaking as follows.
\begin{align*}
\mathcal{L}(\Pi, q, \hat p) = \sum_{(i, A, s) \in \B} q_s \cdot \log\bigg( \frac{\hat p_i}{\sum_{k\in A} \hat p_k} \bigg) \,.
\end{align*}
Focusing on a single index $i \in [n]$, the terms in the weighted log-likelihood that depends on $\hat p_i$ is
\begin{align*}
\mathcal{L}_i(\Pi, q,\hat p) = \sum_{(j, A, s) \in \B: i = j} q_s \cdot \log\bigg( \frac{\hat p_j}{\sum_{k\in A}\hat p_k} \bigg) + \sum_{(j, A, s) \in \B: i \neq j, i\in A} q_s \cdot \log\bigg( \frac{\hat p_j}{\sum_{k\in A} \hat p_k} \bigg) \,.
\end{align*}
Taking the derivative of this function with respect to $\hat p_i$ and setting the result to $0$ gives
\begin{align*}
\frac{\partial \mathcal{L}_i(\Pi, q, \hat p)}{\partial\hat p_i} = \sum_{(j, A, s) \in \B: i = j} q_s \cdot \bigg(\frac{1}{\hat p_i} - \frac{1}{\sum_{k\in A} \hat p_k} \bigg) + \sum_{(j, A, s) \in \B: i \neq j, i\in A} \bigg( -\frac{q_s}{\sum_{k\in A} \hat p_k} \bigg) = 0\,.
\end{align*}
Multiplying both sides by $\hat p_i$ gives
\begin{align*}
\sum_{(j, A, s) \in \B: i = j} q_s \cdot \bigg(1- \frac{\hat p_i}{\sum_{k\in A} \hat p_k} \bigg) + \sum_{(j, A, s) \in \B: i \neq j, i\in A} \bigg( -\frac{q_s \hat p_i}{\sum_{k\in A} \hat p_k} \bigg) = 0\,.
\end{align*}
Rearranging the terms gives
\begin{align*}
\sum_{(j, A, s) \in \B: i = j} q_s \cdot \sum_{u\in A, u\neq i} \frac{\hat p_u}{\sum_{k\in A}\hat p_k} = \sum_{(j, A, s) \in \B: i \neq j, i\in A} \bigg(\frac{q_s \hat p_i}{\sum_{k\in A} \hat  p_k} \bigg) \,.
\end{align*}
The reader may recognize that this equality is essentially
\begin{align*}
\sum_{j\neq i}  \hat p_j \, \underbrace{\sum_{(i,A, s) \in \B_{i\succ j} }  \frac{q_s}{\sum_{k\in A}  \hat p_k}}_{M(\hat p)_{ji}} = \sum_{j\neq i}p_i \,\underbrace{\sum_{(j,A, s) \in \B_{j\succ i} } \frac{q_s}{\sum_{k\in A}  \hat p_k}}_{M(\hat p)_{ij}} \,,
\end{align*}
where $M(\hat p)_{ij}$ is defined in Equation (\ref{eqn:markov-chain}). This is the stationarity condition for a Markov chain with stationary distribution $\hat p$ and state transition probabilities $M(\hat p)_{ij}$'s. This system of equation also means that the weighted MLE $\hat p$ satisfies a notion of self-consistency. Namely,
$$ \hat p^\top M(\hat p) = \hat p\,. $$
If we consider the iterative process of Algorithm \ref{alg:weighted-luce}, then $\hat p$ is a fixed point iterate. The rest of this section is devoted to proving that, starting from any initial distribution, the process converges to the weighted MLE $\hat p$.

\begin{reptheorem}{thm:weighted-lsr}  The output of weighted LSR (Algorithm \ref{alg:weighted-luce}) is the maximum weighted log-likelihood estimate:
$$ \theta_q^{\text{MLE}} = \arg\max_{\theta} \sum_{l=1}^m \, q_l \cdot \log \Pr^{\text{PL}}(\pi_l \,\lvert\, \theta ) \,, $$
where $ \Pr^{PL}(\pi_l\,\lvert \theta )$ is shown in Equation (\ref{eqn:plackett-luce-likelihood}).
\end{reptheorem}

\begin{proof} We adapt our proof from the proof of Theorem 1 in \cite{maystre2015fast}. We first prove a result analogous to Lemma 1 in that paper. Firstly, for a integer $t \geq 1$, define
$$ T(p^\top) = p^\top M(p) $$
and
$$ S_t(p^\top) = \underbrace{T(T( \ldots T(p)))}_{t \text{ times}} \,. $$
As shown previously, $\hat p$ satisfies ${\hat p}^\top = T({\hat p}^\top)$. That is, $\hat p$ is a fixed point of the mapping $T$. If we can also show that $T$ is a \emph{contraction mapping}, i.e.,  $\lVert T(x) - T(y) \rVert \leq c \cdot \lVert x - y\rVert \,\forall x, y \in \calP_n$ for some $c < 1$, then we could guarantee that 
$$ \lim_{t\rightarrow \infty} S_t(p^\top) = {\hat p}^{\top} \quad \forall p \in \calP_n \,. $$
Towards this end, we need to analyze the Jacobian $J$ of the transformation $T$. Firstly, note that $J \in \R^{{n\times n}}$. Recall that
\begin{equation*}
M(p)_{ij} = \begin{cases} \epsilon \cdot \sum_{(i,A,s) \in \B_{j\succ i}} \,\frac{w_s}{\big(\sum_{k\in A} p_k\big) } &\text{ if } i\neq j \\
1 - \epsilon \cdot \sum_{(i,A,s) \in \B_{j\succ i}} \,\frac{w_s}{\big(\sum_{k\in A} p_k\big) } &\text{ otherwise} \\
\end{cases}\,.
\end{equation*}
By definition of the Jacobian, we have for an entry $i, j$ where $i\neq j$ (an off diagonal entry),
$$ [J(p^\top)]_{ij} = \bigg(\frac{\partial p^\top M(p)}{\partial p_i}\bigg)_j = \frac{\partial \big(p^\top M(p)\big)_j }{\partial p_i} = \frac{\partial \big(\sum_k p_k M(p)_{kj} \big) }{\partial p_i} \,. $$
We will now expand on this quantity.
\begin{align*}
[J(p^\top)]_{ij} &=  \frac{\partial \big(\sum_k p_k M(p)_{kj} \big) }{\partial p_i}\\
&= \frac{\partial p_i M(p)_{ii}}{\partial p_i} + \sum_{k\neq i} \frac{\partial \big(p_k M(p)_{kj} \big)}{\partial p_i}\\
&= M(p)_{ij} + p_i \cdot \frac{\partial P_{ij}}{\partial p_i} + \sum_{k\neq i} \frac{\partial (p_k M(p)_{kj}) }{\partial p_i}\\
&= M(p)_{ij} + p_i \cdot \frac{\partial P_{ij}}{\partial p_i} + \frac{\partial (p_j M(p)_{jj}) }{\partial p_i} + \sum_{k\neq i, j} \frac{\partial (p_k M(p)_{kj}) }{\partial p_i}\\
&= M(p)_{ij} + \frac{\partial (p_j M(p)_{jj}) }{\partial p_i} + \sum_{k\neq j} p_k \cdot \frac{\partial M(p)_{kj} }{\partial p_i} \quad (*) \,.\\
\end{align*}
We will expand on the second and the third term.
\begin{align*}
&\frac{\partial (p_j M(p)_{jj}) }{\partial p_i} + \sum_{k\neq j} p_k \cdot \frac{\partial M(p)_{kj} }{\partial p_i}\\
&= \sum_{k\neq j} p_k \cdot \epsilon \cdot \sum_{j, A, s\in \B_{j\succ k, i}} \,\frac{-w_s}{(\sum_{k' \in A} p_{k'})^2} - p_j \cdot \epsilon \cdot \sum_{k\neq j} \frac{\partial M(p)_{jk}}{\partial p_i}\\
&= \epsilon \cdot \sum_{j, A, s\in \B_{j\succ i}} \sum_{k\in A, k \neq j} \,\frac{-w_sp_k}{(\sum_{k' \in A} p_{k'})^2} + \epsilon \cdot \sum_{k\neq j} \sum_{k, A, s \in \B_{k\succ j, i}} \frac{w_s p_j}{(\sum_{k' \in A} p_{k'})^2 }\\
&= -\epsilon \cdot \sum_{j, A, s\in \B_{j\succ i}} \sum_{k\in A, k \neq j} \, w_s \cdot \bigg(\frac{1}{ \sum_{k' \in A} p_{k'}} - \frac{p_j}{(\sum_{k' \in A} p_{k'})^2}\bigg) + \epsilon \cdot \sum_{k\neq j} \sum_{k, A, s \in \B_{k\succ j, i}} \frac{w_s p_j}{(\sum_{k' \in A} p_{k'})^2 }\\
&= - \underbrace{\epsilon \cdot \sum_{j, A, s\in \B_{j\succ i}} \, \frac{w_s}{ \sum_{k' \in A} p_{k'}}}_{M_{ij}} + \epsilon \cdot \sum_{j, A, s\in \B_{j\succ i}} \, \frac{w_sp_j}{(\sum_{k' \in A} p_{k'})^2} +  \epsilon \cdot \sum_{k\neq j} \sum_{k, A, s \in \B_{k\succ j, i}} \frac{w_s p_j}{(\sum_{k' \in A} p_{k'})^2 }\\
&= M(p)_{ij} + \epsilon \cdot \sum_{k} \sum_{k, A, s \in \B_{k\succ j, i}} \frac{w_s p_j}{(\sum_{k' \in A} p_{k'})^2 }\,,
\end{align*}
where, as an overload of notations, $\B_{j\succ j, i} = \B_{j\succ i}$. In the last step, we have combined the last two terms into one.
Subtituting the above into the display $(*)$ gives
$$ [J(p^\top)]_{ij} = \epsilon \cdot \sum_{k} \sum_{k, A, s \in \B_{k\succ j, i}} \frac{w_s p_j}{(\sum_{k' \in A} p_{k'})^2 }\,. $$

Recall that this is just for the off-diagonal terms of the Jacobian. For the diagonal terms, note that
\begin{align*}
\sum_{j=1}^n [J(p)]_{ij} &= \sum_{j} M(p)_{ij} + \sum_{j}\sum_k \,p_k \cdot \frac{\partial M(p)_{kj}}{\partial p_i}\\
&=1 + \sum_k \,p_k \sum_j \, \underbrace{\frac{\partial}{\partial p_i} \bigg( \sum_j M(p)_{kj} \bigg)}_{= 0} = 1
\end{align*}

This means that the Jacobian matrix is a row-stochastic matrix (its rows sum to 1). Now recall the operator $S_t$ defined earlier as the successive application of the operator $T$ for $t$ times. Following the same argument as in the proof of Theorem 1 of \cite{maystre2015fast}, for any initial distribution $p$, so long as $p$ lies within the open probability simplex (its entries are strictly positive) there exists an interger $t_0$ such that $S_t(p)$ is entrywise positive for $t \geq t_0$. Note that by definition,
$$ S_t'(p) = (T^t(p))' = \prod_{o=0}^{t-1} T'(S_o(p))\,. $$
Since $T' = J$ has been shown to be a row stochastic matrix and $S_o(p)$ is entrywise positive, $S_t'(p)$ is also entrywise positive and the rows of $S_t'(p)$ all sum to 1. The rest of the proof follows identically to the proof of Theorem 1 in \cite{maystre2015fast}.
% This means that there exists some $\delta > 0$ such that
% $$ S_t'(p) = \delta \,\mb 1_{n\times n} + R(p) $$
% where $R(p)$ is some matrix satisfying $\lVert R(p)\rVert_1 = 1 - n\delta = c < 1$.
% Consider any $p \neq \hat p \in \calP_n$. Let 
\end{proof}

\newpage
\section{Extra Experiments Discussions}

~\\
\textbf{Datasets Metadata.} Table \ref{tbl:datasets-metadata} summarizes the metadata for all the real-life datasets used in our experiments.
\begin{table}[h!]
\centering
\begin{tabular}{|c|c|c|c|}
\hline
Dataset     & $m$ & $n$ & Reference \\ \hline
APA      & 18723 & 5 & \cite{diaconis1989generalization}  \\ \hline
Irish-Meath & 64081 & 14 &\cite{mattei2013preflib} \\ \hline
Irish-North & 43942 & 12 &\cite{mattei2013preflib} \\ \hline
Irish-West & 29988 & 9 &\cite{mattei2013preflib} \\ \hline
SUSHI & 4997 & 10 &\cite{kamishima2003nantonac} \\ \hline
ML-10M      & 71567 & 10681 & \cite{harper2015movielens}  \\ \hline
\end{tabular}
\caption{Datasets metadata and references. \label{tbl:datasets-metadata}}
\end{table}

~\\
\textbf{Data Pre-processing.} For the three Irish election datasets that contains ties, we perform random tie breaking. The APA and SUSHI datasets contain no ties. For the ML-10M dataset, we first construct a user-item rating matrix from the ratings data. We then apply non-negative low rank matrix completion algorithm (with rank 30) using the open-source implementation found in \cite{Zitnik2012} to `fill in' the missing entries. For each user, the ranking for the items is determined by the value of the filled in entries. To select the subset of items for use in the experiments, we select the items with the highest variance in user rankings. For the small scale real life datasets, we use all the available training rankings and Table \ref{tbl:small-datasets} shows the performance for the algorithms when given all available training data. For variants of the ML-10M datasets, we use up to 14k rankings for training. For model selection, $K$ is chosen from $[2,3,4,5,6,7,8,9,10]$ using Bayesian Information Criterion on the validation set. BIC is defined as $d \log(m_{\text{val}}) - 2\log(L)$ where $d$ is the number of parameters in the mixture model, $m_{\text{val}}$ is the number of validation rankings and $L$ is the likelihood evaluated on the validation set.

~\\
\textbf{Some Notes on Implementations.} We adopt the implementation of the MM algorithm from \citet{choix}. To solve the optimization problems in the M-step in EM-CML and EM-GMM, we use scipy's built-in SLSQP solver. Please refer to our code in the supplementary materials for the implementation of all algorithms used in our experiments. The final class probability estimates are taken as $\hat \beta^k \propto \sum_{l=1}^m q_l^k$ where $q_l^k$ is the posterior class probability computed using the final estimate $\hat\btheta$.

~\\
\textbf{Partial rankings.} Firstly, note that the EM-LSR algorithm proposed in our paper can be decomposed into two steps. The first is an initialization procedure and the second is an EM refinement procedure. In the case of full rankings, initialization is delegated to a clustering-then-estimate approach where the clustering algorithm is spectral clustering. For partial rankings, we can replace spectral clustering by kernel clustering using semi-definite programming (SDP) \cite{peng2007approximating}. We can use either the Kendall kernel or the Mallows kernel \cite{jiao2015kendall} and their extensions to handle partial rankings \cite{jiao2015kendall,lomeli2018antithetic}. Once we have partitioned the rankings into clusters, the remaining of the spectral initialization proceeds in the same way. One can estimate the pairwise preference probabilities for each cluster and apply the least squares parameter estimation algorithm (Algorithm \ref{alg:least-squares-estimate}). 

For the EM-procedure, note that the E-step can still be done efficiently because the PL likelihood for partial rankings can still be evaluated in closed form (see below). For the M-step, the weighted LSR algorithm can also handle partial rankings as it takes advantage of the decomposability of the Plackett-Luce likelihood function. The intuition for this comes from the interpretation of the generative process behind PL as a sequence of choice. Consider a partial ranking $\pi$, of $s$ items where $s < n$, the universe size. The likelihood function for partial ranking is given as
$$ \Pr(\pi \,\lvert\, \theta ) = \frac{e^{\theta_{\pi_1}}}{\sum_{j\geq 1} \theta_{\pi_j} }\ldots \frac{e^{\theta_{\pi_{s-1}}}}{e^{\theta_{\pi_{s-1}}} + e^{\theta_{\pi_{s}}}}\,.$$
Following the same construction as described in Section \ref{sect:algorithm}, one can perform choice breaking on partial rankings just like for full rankings and the algorithm requires little modifications. One can also inspect the proof of Theore \ref{thm:weighted-lsr} and see that the log-likelihood function decomposes into individual choice enumerations. Therefore, Theorem \ref{thm:weighted-lsr} also applies to the partial rankings setting as well.

~\\
\textbf{Ties.} As an example, we might observe a ranking 
$$\pi = [\pi_1, \pi_2, \{\pi_3, \ldots, \pi_u \}, \pi_{u+1}, \ldots, \pi_s]\,,$$
where the items $\{\pi_3, \ldots, \pi_u\}$ are tied. To handle ties, we consider the size of the set of the tied items.
If the set of tied items is small, we generate multiple copies of the original ranking, each with a possible permutation among the tied items, and divide the ranking weight by the number of such permutations. If the set of tied items is large, repeat this process but with a constant number of random permutations from the tied items. Alternatively we can also use a sampling algorithm recently proposed by \cite{liu2019learning}. This sampling algorithm draws linear extensions (full rankings) consistent with given partial rankings and specified model parameters.

\subsection{Additional Experiments}
\subsubsection{Random initialization vs spectral initialization}
As mentioned in our main paper, we observe that spectral initialization outperforms random initialization. To help the reader see more clearly how much initialization helps, we augment Figure 1 in the main paper with results for random initialization and spectral initialization. The figure below shows how a random guess and the spectral initialization algorithm (no EM refinement) perform against the algorithms in the paper. We can see that the EM refinement step does result in a more accurate final estimate than spectral initialization alone. More importantly, spectral initialization provides a significantly better initial estimate than random initialization.

\begin{figure}[h!]
    \centering
    \begin{subfigure}[h]{0.45\textwidth}
        \centering
        \includegraphics[scale=0.4]{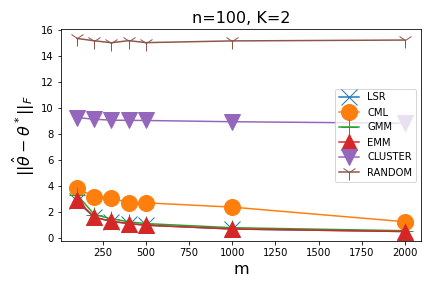}
    \end{subfigure}%

    \begin{subfigure}[h]{0.45\textwidth}
        \centering
        \includegraphics[scale=0.4]{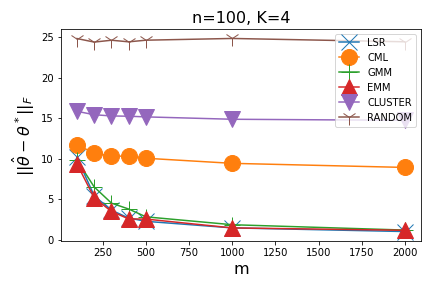}
    \end{subfigure}

    \begin{subfigure}[h]{0.45\textwidth}
        \centering
        \includegraphics[scale=0.4]{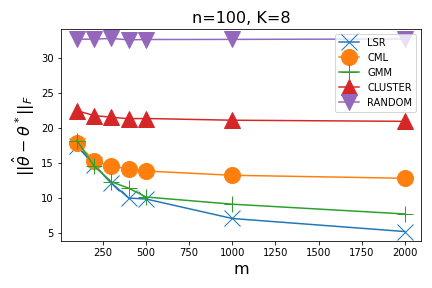}
    \end{subfigure}

    % \begin{subfigure}[h]{0.45\textwidth}
    %     \centering
    %     \includegraphics[scale=0.4]{figs/syn_100_8_ell2_full.png}
    % \end{subfigure}

    \caption{\textbf{Synthetic datasets:} The EM refinement procedure results in more accurate estimate than spectral initialization alone. Spectral initialization outperforms random initialization. \label{fig:initialization}}
\end{figure}

\subsubsection{Comparison to other methods}
\textbf{Bayesian Methods.} In the main paper we focus on parameter estimation and frequentist algorithms. Besides the frequentist (maximum likelihood) approach, some authors have also proposed Bayesian algorithms for PL parameter inference \cite{guiver2009bayesian,caron2012efficient} for the classical PL model. For a mixture of PL models, we are only aware of the works by \citet{mollica2017bayesian} and \citet{caron2014bayesian}. The latter focus on clustering mixtures of Plackett Luce models while the former presented a learning algorithms for a finite mixture of PLs. Here we include extra experiments comparing EM-LSR and this Bayesian algorithm. To obtain final parameter estimates from the Bayesian model, we sample a 1000 samples from the posterior distribution and compute the average. \textbf{Smooth vs Stochastic M-step.} As noted in our main paper, \cite{liu2019learning} used the unweighted LSR algorithm in their EM algorithm for learning mixtures of PL models. In order to incorporate unweighted LSR, they perform a probabilistic clustering in the M-step. The class posterior probabilities computed in the E-step are used to perform the clustering. 

\begin{figure}[]
    \centering
    \begin{subfigure}[h]{0.45\textwidth}
        \centering
        \includegraphics[scale=0.4]{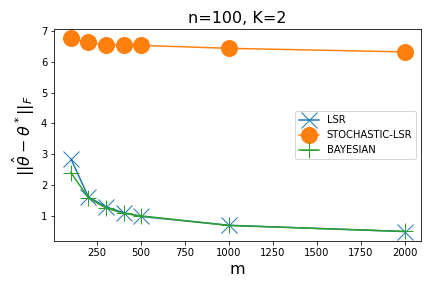}
    \end{subfigure}%
        \hspace{0.5cm}
    \begin{subfigure}[h]{0.45\textwidth}
        \centering
        \includegraphics[scale=0.4]{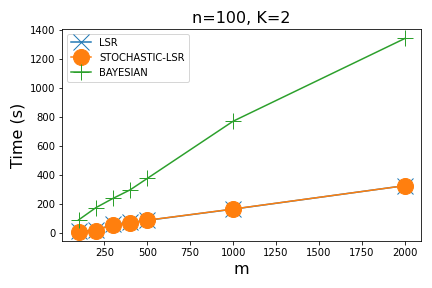}
    \end{subfigure}%

    \begin{subfigure}[h]{0.45\textwidth}
        \centering
        \includegraphics[scale=0.4]{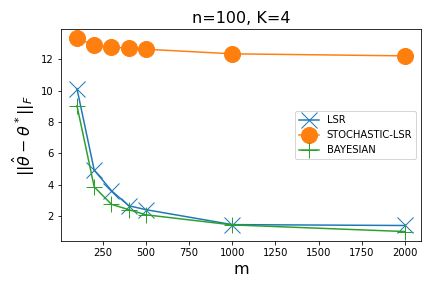}
    \end{subfigure}
        \hspace{0.5cm}
    \begin{subfigure}[h]{0.45\textwidth}
        \centering
        \includegraphics[scale=0.4]{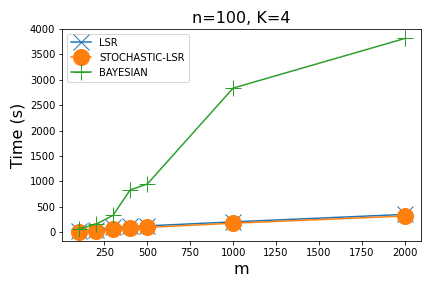}
    \end{subfigure}

    \begin{subfigure}[h]{0.45\textwidth}
        \centering
        \includegraphics[scale=0.4]{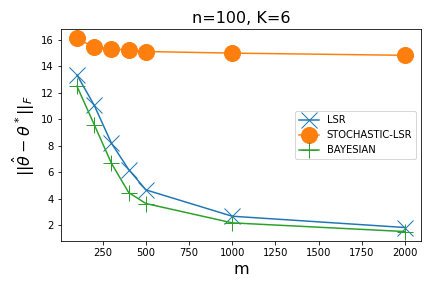}
    \end{subfigure}
        \hspace{0.5cm}
    \begin{subfigure}[h]{0.45\textwidth}
        \centering
        \includegraphics[scale=0.4]{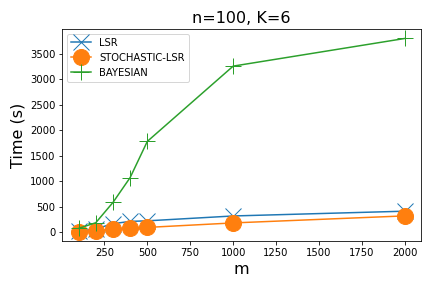}
    \end{subfigure}

    \begin{subfigure}[h]{0.45\textwidth}
        \centering
        \includegraphics[scale=0.4]{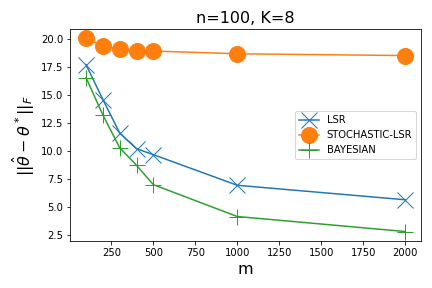}
    \end{subfigure}
        \hspace{0.5cm}
    \begin{subfigure}[h]{0.45\textwidth}
        \centering
        \includegraphics[scale=0.4]{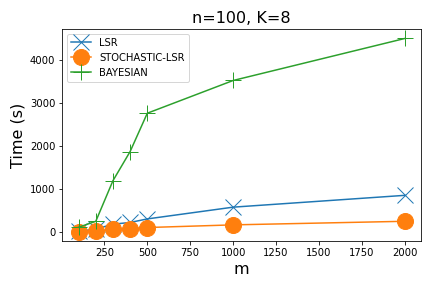}
    \end{subfigure}
    \caption{\textbf{Synthetic datasets:} The Bayesian method produces slightly better estimates than smooth EM-LSR but is much slower than the EM-LSR algorithms. Smooth EM-LSR is significantly more accurate than stochastic EM-LSR. \label{fig:synthetic-extra}}
\end{figure}

\begin{figure}[]
    \centering
    \begin{subfigure}[h]{0.45\textwidth}
        \centering
        \includegraphics[scale=0.4]{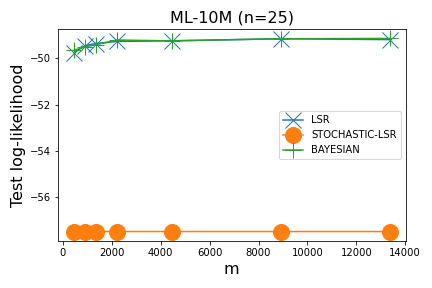}
    \end{subfigure}%
        \hspace{0.5cm}
    \begin{subfigure}[h]{0.45\textwidth}
        \centering
        \includegraphics[scale=0.4]{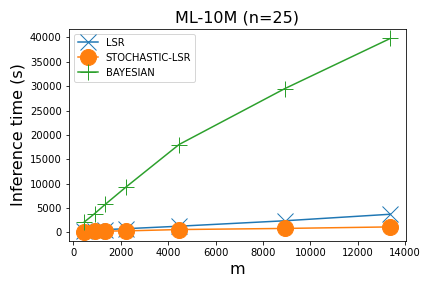}
    \end{subfigure}%

    \begin{subfigure}[h]{0.45\textwidth}
        \centering
        \includegraphics[scale=0.4]{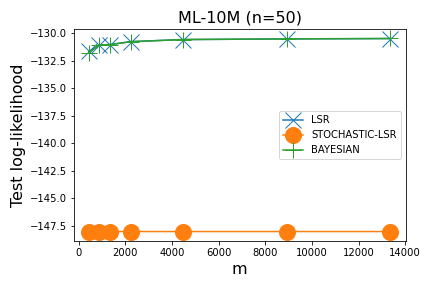}
    \end{subfigure}
        \hspace{0.5cm}
    \begin{subfigure}[h]{0.45\textwidth}
        \centering
        \includegraphics[scale=0.4]{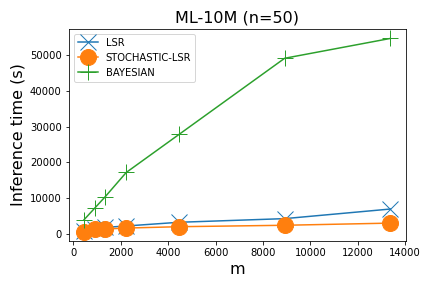}
    \end{subfigure}

    \begin{subfigure}[h]{0.45\textwidth}
        \centering
        \includegraphics[scale=0.4]{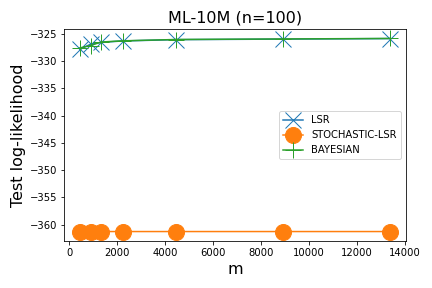}
    \end{subfigure}
        \hspace{0.5cm}
    \begin{subfigure}[h]{0.45\textwidth}
        \centering
        \includegraphics[scale=0.4]{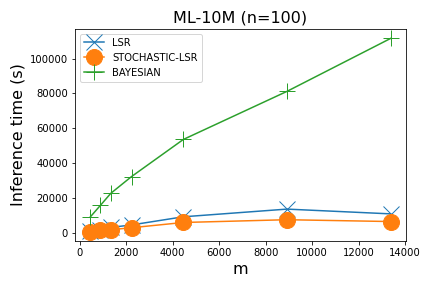}
    \end{subfigure}

    \caption{\textbf{ML-10M datasets:} On real datasets, soft EM-LSR is more accurate than stochastic EM-LSR. Smooth EM-LSR is similar in accuracy to the Bayesian method but is orders of magnitude faster.\label{fig:real-extra}}
\end{figure}

We perform additional experiments on synthetic datasets and ML-10M datasets comparing these three algorithms. Figure \ref{fig:synthetic-extra} shows the results on synthetic datasets and Figure \ref{fig:real-extra} shows the results on ML-10M datasets. EM-LSR and the Bayesian algorithm have their own advantages over the other algorithm. EM-LSR is faster and requires little tuning. The Bayesian method, on the other hand, can be used to obtain uncertainty quantification as we learn a posterior distribution over the parameters $\btheta$ and produces slightly better estimates, especially in the data-poor regime. The tradeoff is of course speed. While the EM-LSR algorithm with a probabilistic clustering M-step is approximately faster by a constant factor than the original EM-LSR algorithm (both are orders of magnitude faster than the Bayesian method), it produces significantly worse estimates. We suspect that the noise introduced by the stochastic M-step causes the trajectory of the EM algorithm to end up in a bad local optimum.

\newpage 
\bibliographystyle{plainnat}
\bibliography{references}